\documentclass[journal]{IEEEtran}
\usepackage{amsmath,amssymb}
\usepackage{booktabs}
\usepackage{cite}
\usepackage{graphicx}
\usepackage{makecell}
\usepackage{multirow}
\usepackage{placeins}
\usepackage{float}
\usepackage{supertabular}
\usepackage{flafter}
\usepackage[linesnumbered,ruled,vlined]{algorithm2e}
\usepackage{tikz}
\usetikzlibrary{arrows.meta,fit,positioning}
\usepackage[hidelinks,bookmarksnumbered,bookmarksopen,bookmarksopenlevel=2]{hyperref}
\definecolor{mygray}{gray}{.9}
\newcommand{\rkii}[1]{\underline{#1}}
\definecolor{myblue}{rgb}{.35,.67,.99}
\makeatletter
\newcounter{subparagraph}[paragraph]

\newcommand{\subparagraph}{\@startsection{subparagraph}{4}{\parindent}%
  {1.5ex plus .5ex minus .2ex}{-1em}{\normalfont\normalsize\itshape}}
\makeatother

\begin{document}
\raggedbottom

\title{Selective-Rerouter: A Rerouting Strategy for Hypothetical Post-Disaster Urban Scenarios with a Thin-Boundary Co-Simulator}

\author{Xiao~Wang, Shun-Ren~Yang,~\IEEEmembership{Member,~IEEE}, and Hui-Nien Hung
\thanks{X. Wang and S.-R. Yang are with the Department of Computer Science and the Institute of Communications Engineering, National Tsing Hua University, Hsinchu 30013, Taiwan (e-mail: sryang@cs.nthu.edu.tw).}
\thanks{H. Hung is with the Institute of Statistics, National Chiao Tung University, Hsinchu, Taiwan (e-mail: hhung@stat.nctu.edu.tw).}
\thanks{The Selective-Rerouter is the intact-side algorithm. Our co-simulator couples it with the DTN-UAV Forwarder, the fixed module inside a damaged corner. Section~\ref{sec:hybrid400} reports one hybrid-cost run on the Tainan road-network topology, with the partition artificially injected.}%
}

\markboth{IEEE Transactions draft}%
{Wang \MakeLowercase{\textit{et al.}}: Selective-Rerouter}

\maketitle

\begin{abstract}
Hypothetical post-disaster urban environments may be partitioned into disjoint delay-tolerant-network (DTN) subregions under a thin-boundary information-isolation constraint, which severely limits cross-zone message delivery and makes centralized rerouting impractical. To address this, we propose Selective-Rerouter, a delayed-binding anonymous rerouting strategy for such partitioned environments. The scheme jointly exploits road occupancy and traffic speed for bottleneck detection, adaptively enlarges its upstream intervention window with congestion severity to include vehicles imminently approaching bottlenecks, and evaluates paths via horizon-aware fusion of live measurements and short-term forecasts, where only the prediction head of LSTAN-GERPE~\cite{WangYang2025LSTAN} is fine-tuned with a path-ranking loss to preserve travel-time ordering, while a driver-aware scaling factor adjusts forecasts without altering live observations. Vehicles reroute only upon bottleneck encounters and keep $K$ candidate routes. To assess performance under information isolation, we build a SUMO-based co-simulator coupling Selective-Rerouter on the intact side with a fixed DTN-UAV Forwarder inside each damaged corner. The forwarder consumes only scalar cost-announcements, performs no path-cost computation, and resolves civilian identities solely after bundles reach gateway buffers, thus acting as a testbed component rather than a novel DTN routing algorithm. Evaluations include standalone validation over undamaged districts, corner-level verification of cost-announcement-compliant bundle handoff, and four-corner tests that replace only the intact-side rerouting policy. Results show that hybrid live-forecast costing yields dominant travel-time gains. All experiments use discrete SUMO runs based on real-world urban road topology with artificially injected network partition damage, with no physical prototype.
\end{abstract}


\section{Introduction}
Hypothetical post‑disaster urban events can severely damage road and communication infrastructure, fragmenting cities into disjoint delay‑tolerant network (DTN) subregions. In such partitioned settings, end‑to‑end multi‑hop inter‑zone connectivity disappears, and cross‑region message delivery must rely on store‑and‑forward mechanisms. We denote each disconnected subregion as a \textit{corner}: no intact road path connects it to the rest of the city, so conventional multi‑hop routing cannot evacuate messages from the subregion. Remaining mutually‑reachable road segments form the \textit{intact network}.

In this paper, each self‑contained end‑to‑end application delivery unit is termed a \textit{bundle}, following standard delay‑tolerant network (DTN) terminology. A bundle may be cached for extended durations and relayed via store‑carry‑forward under intermittent connectivity. Emergency operations face two concurrent objectives: sustaining civilian mobility over intact roads and delivering bundles across isolated corners.

Two largely disjoint research lines address partial aspects of this problem. Selective rerouting optimizes routing for fully‑connected intact road graphs with full access to civilian‑vehicle identities. By contrast, DTN forwarding and UAV‑assisted relays enable bundle delivery within disconnected subregions, yet cannot reuse civilian travel trajectories computed on the intact‑network side. Information exchange between intact zones and damaged corners is restricted to a \textit{thin boundary}, consisting of intact edges adjacent to damaged subregions. Every gateway of a damaged corner lies on this boundary and is an ingress‑egress point. No gateway is placed inside the corner. A hard constraint prohibits transmitting civilian‑vehicle identifiers across the thin boundary. Accordingly, intact‑side rerouting modules shall publish a \emph{gateway-to-gateway cost-announcement} (\emph{cost-announcement}), a scalar annotated only with the gateway index and carrying no civilian-vehicle identifier. Civilian carriers may be assigned to bundles \textit{only after} bundles physically arrive at gateway buffers.

Section~\ref{sec:related} leaves three gaps for this setting. Selective rerouting is built for a fully connected road network and does not publish a cost-announcement or defer civilian binding until a bundle reaches a gateway buffer. UAV-assisted DTN forwarding stays inside the disconnected region and does not use civilian trips on the intact network as carriers. The two lines are also evaluated separately, with no co-simulation that enforces the thin-boundary constraint and holds the corner forwarder fixed. On a fully connected network the same three-phase pipeline still mis-detects congestion, selects the wrong vehicles, and mis-orders paths over a forecast horizon (Section~\ref{sec:challenges}).

To close these gaps, this paper presents Selective-Rerouter, a deferred-binding anonymous rerouting strategy for DTN-partitioned hypothetical post-disaster urban environments. It publishes cost-announcements and names a civilian carrier only after the bundle arrives at a gateway buffer, so an intact-network trip can carry the bundle without sending a vehicle identifier across the thin boundary. On the intact network it uses joint occupancy-speed bottleneck detection, a severity-adaptive upstream window that includes vehicles imminently approaching congestion, and horizon-aware path costs that fuse live observations with short-term forecasts. Only the prediction head of LSTAN-GERPE~\cite{WangYang2025LSTAN} is fine-tuned with a path-ranking loss, which preserves travel-time order among candidate routes, and a driver-aware scaling factor adjusts forecasts without changing live measurements. Civilian trips keep multiple candidate routes. For joint evaluation we build a SUMO-based co-simulator. The DTN-UAV Forwarder stays fixed inside each corner, consumes cost-announcements, and is a testbed module rather than a new DTN routing algorithm (Section~\ref{sec:dtn}). Comparisons replace only the intact-side rerouter.
\label{sec:intact‑motivation}

The main contributions of this work are summarized as follows:
\begin{enumerate}
\item \textbf{Selective‑Rerouter}: An intact‑side anonymous rerouting policy implementing deferred bundle‑to‑civilian binding, where civilian carriers are assigned exclusively after bundles reach gateway buffers (Section~\ref{sec:selch}). In benchmark comparisons, only the rerouting primitive is substituted. We report quantitative metrics including $\bar{t}_{\mathrm{civ}}$, $\bar{T}_{\mathrm{obj}}$, $T_{\mathrm{obj}}^{\max}$, $D_{\mathrm{osc}}$, and $D_{\mathrm{osc}}/M$ against state‑of‑the‑art baselines: NRR, ReFOCUS+, FCR, CCRGS, Du‑GAQ, and CAIE‑TT‑Scoped.
\item \textbf{Thin-boundary constrained co-simulator}: A SUMO-based co-simulator built around the thin-boundary cost-announcement interface $c_{g,u}$ (Section~\ref{sec:interface}). It enables joint performance assessment of intact-side rerouting and intra-corner forwarding, with the DTN-UAV Forwarder held fixed across all comparative runs. Note that corner-local intermediate statistics reported in Section~\ref{sec:sub1val} are not treated as end-to-end evaluation metrics.
\end{enumerate}

The remainder of this paper is organized as follows. Section~\ref{sec:related} reviews existing literature on selective rerouting and post‑disaster DTN communications. Section~\ref{sec:model} formalizes the hypothetical partitioned urban scenario, intact‑network travel‑cost model, cross‑boundary interface, and the DTN‑UAV Forwarder. Section~\ref{sec:problem} defines the optimization objective, practical challenges, feasibility constraints, and problem scope. Section~\ref{sec:selch} elaborates the Selective‑Rerouter algorithm. Section~\ref{sec:exp} describes the evaluation protocol and presents experimental results. Finally, Section~\ref{sec:conclusion} concludes this paper.

\section{Related Work}
\label{sec:related}
This paper lies at the intersection of two largely separate research communities:
\textit{selective vehicle rerouting} for fully‑connected urban road networks, and
\textit{UAV‑assisted delay‑tolerant networking} for disconnected disaster‑stricken regions.
The target partitioned post‑disaster urban scenario imposes unique thin‑boundary information‑isolation constraints that cannot be fully addressed by either research line in isolation.
Section~\ref{subsec:rerouting_related} reviews existing selective rerouting approaches designed for intact road infrastructure.
Section~\ref{sec:related_uav} surveys UAV‑aided DTN forwarding solutions for disconnected sub‑regions.
Deferred binding, formalized in Section~\ref{sec:boundary}, means a civilian carrier is named only after the bundle reaches a gateway buffer. Compliance of existing designs with thin-boundary isolation and with that rule is examined in Section~\ref{sec:eligibility}.
At the end of this section, we summarize cross‑domain research gaps that motivate our Selective-Rerouter and the accompanying co-simulator.

\subsection{Selective Rerouting}
\label{subsec:rerouting_related}
Microscopic rerouting modifies a subset of vehicle routes to reduce individual or system‑wide travel time \cite{lopez2018sumo,dlr71460,krajzewicz2012recent}.
A canonical rerouting pipeline consists of three sequential phases: congestion detection, selection of vehicles eligible for replanning, and allocation of alternative paths.
Learning‑enabled rerouting solutions differ primarily in where learned components are embedded within this three‑phase workflow.

\subsubsection{Type I: End‑to‑end learned route‑decision rerouters}
Type‑I frameworks employ learned models to directly produce route decisions \cite{Du2024CACAIE,9367130,9516945,9686571,7949069}.
Some fog‑cloud‑based approaches score congested regions and select short alternative paths from learned representations \cite{Du2024CACAIE}.
Other formulations cast traffic regulation as a sequential decision‑making task and output next‑hop directions or speed advisories directly from trained policies \cite{9367130,9516945,9686571,7949069}.

Performance gains within this family stem mainly from end‑to‑end training. These designs rarely adopt explicit congestion validation, controlled rerouting‑vehicle cardinality, or explicit fusion of live measurements and traffic forecasts. Furthermore, end-to-end learned policies lack dedicated modules for cost-announcement generation and deferred civilian-vehicle assignment upon bundle arrival at a gateway buffer. Consequently, Type‑I rerouters target fully‑connected road networks and cannot be directly applied to our partitioned post‑disaster scenario.

\subsubsection{Type II: Pipeline‑preserving rerouters with forecast‑augmented cost}
Type‑II approaches retain the full three‑phase rerouting pipeline and leverage traffic forecasts only as additional inputs for path‑cost evaluation \cite{7243355,7478132,7514969,WangYang2025LSTAN}.
Cloud‑assisted navigation and en‑route guidance inject forecast information into path‑computation procedures \cite{7243355,7478132}.
Other works exploit bottleneck forecasts to trigger or prioritize replanning events \cite{7514969}, or enable dynamic rerouting using velocity predictions \cite{WangYang2025LSTAN}.
The LSTAN‑GERPE model falls into this category: it forecasts segment‑level traffic speed but performs no vehicle selection or route assignment \cite{WangYang2025LSTAN}.
Our Selective‑Rerouter belongs to the Type‑II family: forecasts feed into path‑cost computation while detection, vehicle‑selection, and route‑allocation logic remain outside the learned module.

\subsubsection{Congestion Detection}
Proactive rerouting requires identifying forming bottlenecks before long downstream queues build up.
One class of schemes detects congestion purely from road‑segment density metrics and triggers rerouting accordingly \cite{pan2012proactive}.
However, density‑only indicators may mark a segment as uncongested even when vehicles traveling on that segment already move slowly.
Distributed detection relying on V2V communication assumes homogeneous per‑vehicle speed observations and limited radio range, such that the same physical segment may receive inconsistent congestion labels from different observers \cite{6737608}.
Fusing segment‑level and zone‑level scores may misclassify short local queues as large‑scale regional events or vice‑versa \cite{8901448}.

Under thin‑boundary information isolation, detectors deployed on intact‑network segments can only observe open intact streets.
Single‑metric or purely local detectors therefore provide insufficient evidence for computing valid cost-announcements.
These practical difficulties are elaborated in Section~\ref{sec:challenges}.
To mitigate this limitation, our design jointly thresholds both occupancy and speed for bottleneck identification.

\subsubsection{Rerouted‑Vehicle Selection}
After bottleneck segments are flagged, rerouting systems must select which vehicles receive updated alternative routes.
Naive strategies that weight road segments by severity and redirect vehicles toward locally inexpensive neighbors may increase overall trip durations and simply shift congestion elsewhere \cite{7363112}.
Selecting a fixed number of vehicles near predicted hotspots risks including vehicles that will never enter congested segments while omitting upstream vehicles that will soon arrive \cite{7514969}.
Predefined spatial intervention windows tied to signal controllers, smart‑light coverage, or fixed hop‑count cutoffs exhibit similar inflexibility \cite{7589067,7433412,8324943}.
On irregular urban street layouts, static hop limits trigger excessive replanning under light congestion and select too few vehicles under heavy congestion \cite{Ho2023CVRS}.
Recent studies replace rigid hop‑based bounds with travel‑time filters and residual‑capacity criteria \cite{Ho2023CVRS}.

Within our partitioned‑city setting, static upstream windows operate only over intact‑network topology. Vehicles just about to enter bottlenecks may be excluded from candidate pools and can never become eligible for cost-announcement-driven bundle transportation.
Our solution makes upstream intervention depth adaptive with measured congestion severity and additionally incorporates vehicles imminently approaching bottlenecks.

\subsubsection{Alternative Route Allocation}
The final pipeline stage assigns alternative candidate paths to selected vehicles.
Unconstrained per‑vehicle route updates can relocate network‑wide congestion, motivating designs that maintain multiple candidate routes and composite cost functions.
Scoring candidates purely using instantaneous observed speed and occupancy ignores future traffic evolution \cite{7433412}.
Entropy‑based ranking built upon fused instantaneous states suffers identical limitations: future segment conditions are evaluated using only current snapshots \cite{8901448}.
Although some k‑shortest‑path pipelines incorporate near‑term flow forecasts for replanning triggering, forecast information is not propagated into segment‑level path costs \cite{Tay2025EAAI}.
For trips lasting multiple decision intervals, path evaluation requires combining live observations with short‑horizon forecasts.
This challenge is further detailed in Section~\ref{sec:challenges}.
Our proposed cost formulation fuses live measurements and multi‑horizon forecasts. Only forecasting heads are fine‑tuned with path‑ranking loss, and scaling factors apply exclusively to forecast outputs without altering raw live observations.

While these rerouting techniques are effective for connected cities, they operate under the assumption of end‑to‑end road connectivity and cannot handle geographically partitioned disaster zones. We next turn to DTN‑based solutions designed for disconnected post‑disaster subregions.

\subsection{Post‑Disaster UAV‑Assisted DTN Networks}
\label{sec:related_uav}
Partitioned disaster‑affected cities constitute challenged networks where stable end‑to‑end communication vanishes and bundles rely on store‑carry‑forward delivery \cite{fall2003dtn}.
Classical DTN forwarding paradigms including epidemic flooding, probabilistic routing and message‑ferry mechanisms are well‑established for intermittently connected networks \cite{vahdat2000epidemic,lindgren2003prophet,burgess2006maxprop,zhao2004ferry}. However, these schemes focus purely on relay logic within disconnected zones and lack support for leveraging civilian vehicle mobility from intact road segments.

A recently proposed post‑disaster rerouting framework leverages UAVs for traffic diversion under partial node reachability and investigates generalization across road segments \cite{WlodarczykSaber2026CAIE}.
Nevertheless, it optimizes connectivity and travel performance within its own modeling boundary. It cannot recruit ordinary civilian vehicle trips planned over intact road segments as inter‑correl bundle carriers. Furthermore, existing literature lacks joint co‑simulation frameworks that couple intact‑side traffic rerouting and corner‑local DTN forwarding. These constraints are further examined in Section~\ref{sec:eligibility}.

\textbf{Summary of Research Gaps.}
Taken together, the two surveyed research threads reveal clear unresolved limitations for partitioned post‑disaster urban scenarios subject to thin‑boundary information isolation.

First, existing selective traffic rerouting algorithms target fully‑connected road infrastructures.
They possess no native support for publishing cost-announcements or deferring civilian‑carrier binding until bundles arrive at gateway buffers, both mandatory for cross‑corner message exchange.

Second, conventional UAV‑assisted DTN and store‑carry‑forward solutions primarily optimize message ferrying inside disconnected subregions.
They fail to exploit civilian mobility on intact road segments as inter‑corner data carriers.

Third and most critically, few existing works explore the \textit{cross‑domain coupling} between intact‑side traffic rerouting and corner‑local DTN forwarding.
Prior studies treat traffic rerouting and DTN message relaying as independent problems, lacking standardized co‑simulation testbeds that strictly enforce thin‑boundary information‑isolation constraints for end‑to‑end quantitative evaluation.

These three interrelated gaps motivate our Selective-Rerouter policy and the accompanying co-simulator elaborated in Sections~\ref{sec:system_model}--\ref{sec:selch}.

\section{System Model}
\label{sec:model}
\begin{figure}[!t]
\centering
\includegraphics[width=\columnwidth]{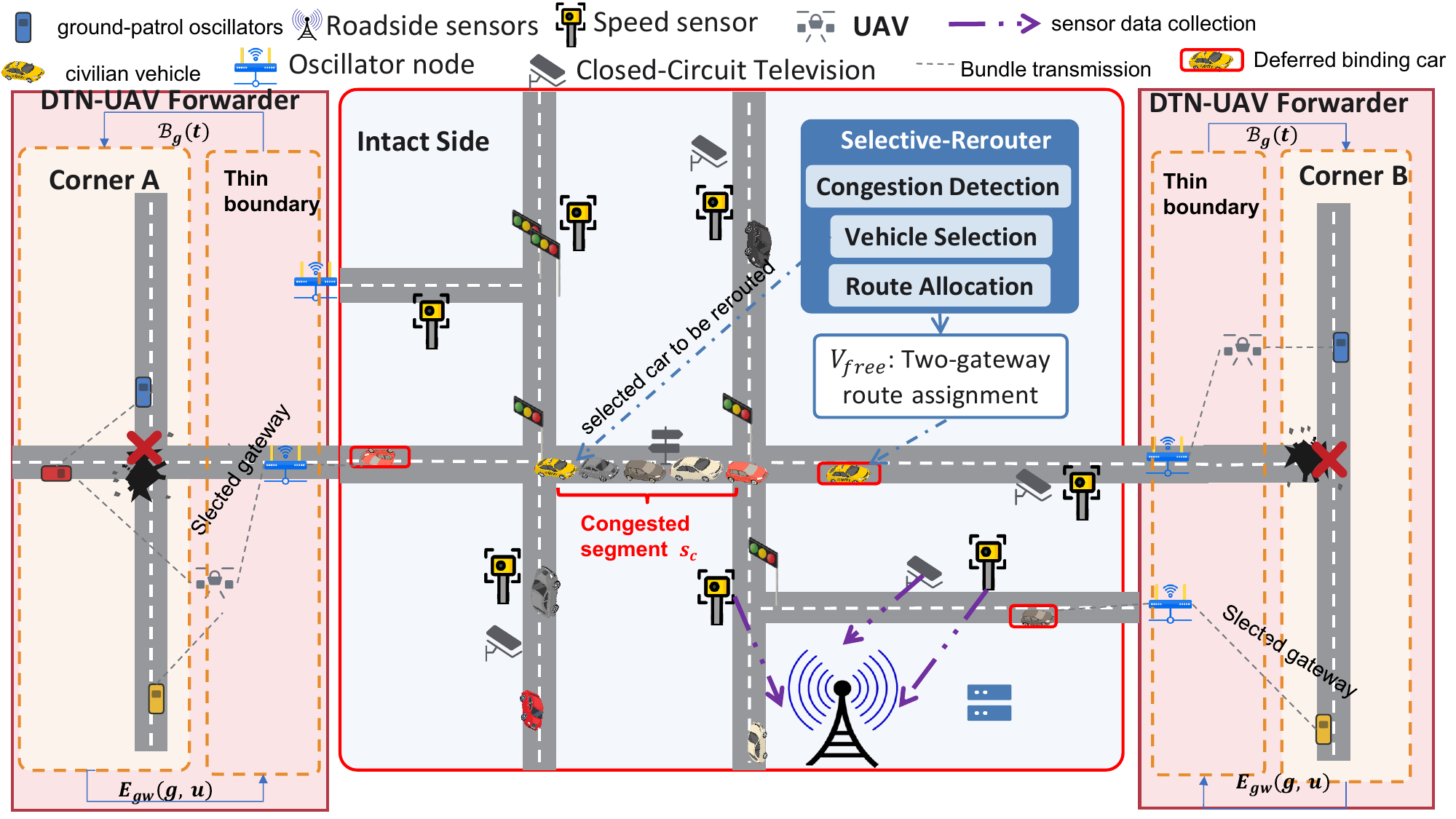}
\caption{Co‑simulation under a unified global clock. Selective‑Rerouter executes on the intact‑network side, while each damaged corner runs a fixed DTN‑UAV Forwarder. Inter‑partition interactions are restricted exclusively to the thin boundary. The beacon $(u,c_{g,u})$ conveys scalar cost values without exposing any vehicle identifier. Civilian carriers are assigned only upon receiving the gateway‑arrival event $(g,u)$. In comparative evaluations, only Selective‑Rerouter is replaced, whereas the DTN‑UAV Forwarder remains unchanged.}
\label{fig:cosim-panels}
\end{figure}

This section details the system model adopted in our co‑simulator. As illustrated in Fig.~\ref{fig:cosim-panels}, the urban road network is partitioned into an \emph{intact network} and multiple disjoint damaged subregions denoted as \emph{corners}, separated by a \emph{thin boundary}.

Our co‑simulator runs two coupled sub‑modules under a unified simulation clock. The \textbf{Selective‑Rerouter} module resides on the intact‑network side, computing rerouting decisions and producing \emph{gateway‑to‑gateway cost‑announcement} outputs. A gateway‑to‑gateway cost‑announcement (shortened to \emph{cost‑announcement}) is a scalar metric published by a gateway, representing the minimal detour overhead incurred when civilian vehicles carry a bundle toward a remote target gateway. Gateways broadcast these scalars within a \emph{beacon}, which consists of entries $\{(u,c_{g,u})\}$ mapping each target‑gateway index $u$ to its corresponding cost. No civilian‑vehicle identifiers are embedded in beacon payloads. Meanwhile, the \textbf{DTN‑UAV Forwarder} operates inside each corner to process bundles via UAVs and ground‑patrol oscillators. In this framework, Selective‑Rerouter generates cost‑announcements, and the DTN‑UAV Forwarder passively consumes these scalar values.

\subsection{Hypothetical Post‑Disaster Scenario and Co‑Simulation Data‑Flow Overview}
\label{sec:scenario}
\label{sec:flow}

Let $\mathcal{Q}$ denote the set of corner subregions with cardinality $|\mathcal{Q}|$. Each corner $q\in\mathcal{Q}$ comprises a gateway set $\mathcal{G}_q$ of size $n_g$, one UAV $a_q$, and a set of ground‑patrol oscillators $\mathcal{O}_q$. All gateways in $\mathcal{G}_q$ are located on the thin boundary between corner $q$ and the intact network, serving as the sole ingress and egress points for bundles and cost‑announcement beacons (see Fig.~\ref{fig:cosim-panels}). Both UAVs and ground‑patrol oscillators are confined within corner boundaries. A \emph{ground‑patrol oscillator} refers to a ground mobile entity that follows predefined, fixed patrol trajectories inside its host corner.

Oscillator trajectories stay static throughout simulation. Civilian vehicles travel over intact road segments as well as thin‑boundary segments. The co‑simulator proceeds in discrete one‑second time steps $t$. Selective‑Rerouter is not triggered at every simulation step; instead, its full rerouting pipeline activates at decision instants $\tau = nT$, where $T$ is the decision period and $n\ge 0$ is an integer. At each activation instant, the pipeline performs congestion detection, eligible‑vehicle selection, and alternative‑route allocation. Every $T_{\mathrm{ref}}$ simulation steps, each gateway recomputes and re‑publishes its cost‑announcements. Vehicle probes fed by roadside sensors and CCTV cameras report vehicle identifiers, current road segments, residual routes, travel destinations, and instantaneous speeds. Rerouting instructions are delivered via downlink as edge sequences.

Gateways and ground‑patrol oscillators within corner $q$ share a common index space $i\in\{0,\dots,n_g-1\}$. The remote gateway index set of corner $q$ is
\begin{equation}
\mathcal{U}_q=\left\{ u \,\big|\, u \text{ is gateway index of corners other than } q\right\},
\label{eq:destset}
\end{equation}
For a bundle $m$, $o^{\star}_m$ denotes the destination ground‑patrol oscillator.

Bundles can leave a corner only after being handed over to a gateway by the DTN‑UAV Forwarder. This handoff is visualized in Fig.~\ref{fig:cosim-panels}.

\subsection{Hybrid Live–Forecast Cost Model for Selective‑Rerouter on the Intact Network}
\label{sec:system_model}
This subsection defines travel‑cost functions over intact road segments for route evaluation and cost‑announcement computation within the Selective‑Rerouter framework.

\subsubsection{Roadside Observation Measurements}
At each decision instant, Selective‑Rerouter consumes multi‑modal sensor measurements collected from roadside infrastructure. Sensors provide vehicle locations, travel destinations, segment‑level road occupancy, and average segment speeds. For each instant $\tau$, the system gathers $N_v(s,\tau)$, $\bar{V}_{\mathrm{o}}(s,\tau)$, and $R_O(s,\tau)$, alongside per‑vehicle status information. Rerouting commands are dispatched only to civilian vehicles whose residual routes intersect congested road segments.

\subsubsection{Dual‑Graph Road Network and Vehicle State}
\label{subsec:graph_vehicle_state}
Let $\mathcal{V}$ be the set of civilian vehicles operating on intact roads. The urban road topology is represented as a dual directed graph $G=(\mathcal{S},\mathcal{A})$. Vertices in $\mathcal{S}$ correspond to intact or thin‑boundary road segments, and directed arcs $\mathcal{A}$ encode valid turning maneuvers. The vertex set $\mathcal{S}$ can be further partitioned into disjoint geographic zones. Each segment $s\in\mathcal{S}$ has static attributes: physical length $L_{rs}(s)$, lane count $N_l(s)$, and speed‑limit $V^{\mathrm{lim}}(s)$. The free‑flow length of path $\pi$ is $\ell(\pi)=\sum_{s\in\pi}L_{rs}(s)$. The scalar $N_{cr}$ denotes the maximum number of candidate alternative routes maintained for each vehicle, and $r_k(v_r,\tau)$ denotes the $k$-th candidate route of vehicle $v_r$ at decision time $\tau$.

The instantaneous state of an in‑traffic vehicle $v$ is given by:
\begin{equation}\label{eq: vehicle_state}
\xi(v,\tau)=\bigl(s(v,\tau),\,d(v),\,r(v,\tau)\bigr),
\end{equation}
where $s(v,\tau)$ is its current road segment, $d(v)$ is its destination, and $r(v,\tau)$ represents its residual replannable route. The complete observable system state is:
\begin{equation}\label{eq: observable_set}
\begin{aligned}
\Omega(\tau)
&=\bigl\{\xi(v,\tau)\bigr\}_{v\in\mathcal{V}_{\mathrm{on}}(\tau)}\\
&\quad\cup\bigl\{N_v(s,\tau),\,\bar{V}_{\mathrm{o}}(s,\tau),\,R_O(s,\tau)\bigr\}_{s\in\mathcal{S}},
\end{aligned}
\end{equation}
where $\mathcal{V}_{\mathrm{on}}(\tau)\subseteq\mathcal{V}$ is the subset of vehicles traveling on intact roads at time $\tau$.

\subsubsection{Congestion Detection Metrics}
Congestion detection is built upon normalized occupancy and normalized speed metrics:
\begin{equation}\label{eq: R_O}
R_O(s,\tau)=\frac{N_v(s,\tau)}{N_v^{\max}(s)},
\end{equation}
\begin{equation}\label{eq: N_v}
N_v^{\max}(s)=\frac{L_{rs}(s)}{\bar{L}_v+L_{sd}}\times N_l(s).
\end{equation}
\begin{equation}\label{eq: R_V}
R_V(s,\tau)=\frac{\bar{V}_{\mathrm{o}}(s,\tau)}{V^{\mathrm{lim}}(s)}.
\end{equation}
We use hard‑trigger thresholds $\delta_O,\delta_V$ and soft‑trigger thresholds $\delta_O^\mathrm{soft},\delta_V^\mathrm{soft}$ for congestion triggering. Their numerical values are reported in Section~\ref{sec:exp}.

\subsubsection{Live Observation and Horizon‑Aware Forecasting}
The driver‑mapping coefficient $Bd(v_r,s)\in[0,1]$ quantifies personalized driving characteristics of vehicle $v_r$ with respect to historical speed statistics on segment $s$. A dimensionless driver‑aware scaling factor $\kappa(v_r,s)$ adjusts forecast speeds to account for driver heterogeneity; $\kappa(v_r,s)$ defaults to $1$ when insufficient historical driving traces for pair $(v_r,s)$ are available.

Let $T_h$ denote the forecast time step, decoupled from rerouting decision period $T$. $\mathcal{F}_{\tau}$ aggregates observable system state $\Omega(\tau)$, historical sensor readings, and static graph features. The fleet‑average forecast speed for road segment $s$ at prediction horizon $h\in[0,T_{\mathrm{out}}-1]$ reads:
\begin{equation}\label{eq:vpred_align}
\bar{V}_{\mathrm{f}}(s,\tau,h)
=\mathbb{E}\Bigl[\bar{V}_{\mathrm{o}}\bigl(s,\tau+(h{+}1)T_h\bigr)\,\Big|\,\mathcal{F}_{\tau}\Bigr].
\end{equation}
This conditional expectation is approximated by the LSTAN‑GERPE model~\cite{WangYang2025LSTAN}. Applying driver‑aware scaling yields personalized forecast speed:
\begin{equation}\label{eq:hatV_def}
\hat{V}_{\mathrm{f}}(s,\tau,h,v_r)=\bar{V}_{\mathrm{f}}(s,\tau,h)\times\kappa(v_r,s).
\end{equation}
A horizon‑dependent blending weight $\alpha(h)\in[0,1]$ decays with increasing prediction horizon to fuse real‑time observations and forecasts:
\begin{equation}\label{eq: hybrid_speed}
\tilde{V}(s,\tau,h,v_r)=\alpha(h)\,\bar{V}_{\mathrm{o}}(s,\tau)+\bigl(1-\alpha(h)\bigr)\,\hat{V}_{\mathrm{f}}(s,\tau,h,v_r).
\end{equation}
Importantly, $\kappa(v_r,s)$ modifies only forecast outputs; raw live sensor measurements $\bar{V}_{\mathrm{o}}(s,\tau)$ remain untouched.

\subsubsection{Segment‑Level Travel‑Time Decomposition}
Segment traversal time consists of signal‑governed queueing delay $t_q$ and free‑flow passing duration $t_p$:
\[
t_v\bigl(s_k^{(j)}(v_r,\tau)\bigr)= t_q\bigl(s_k^{(j)}(v_r,\tau)\bigr)+t_p\bigl(s_k^{(j)}(v_r,\tau)\bigr),
\]
where $t_q$ is computed following the traffic‑signal model from~\cite{8641378}.
\begin{figure}[H]
\centering
\includegraphics[width=\columnwidth]{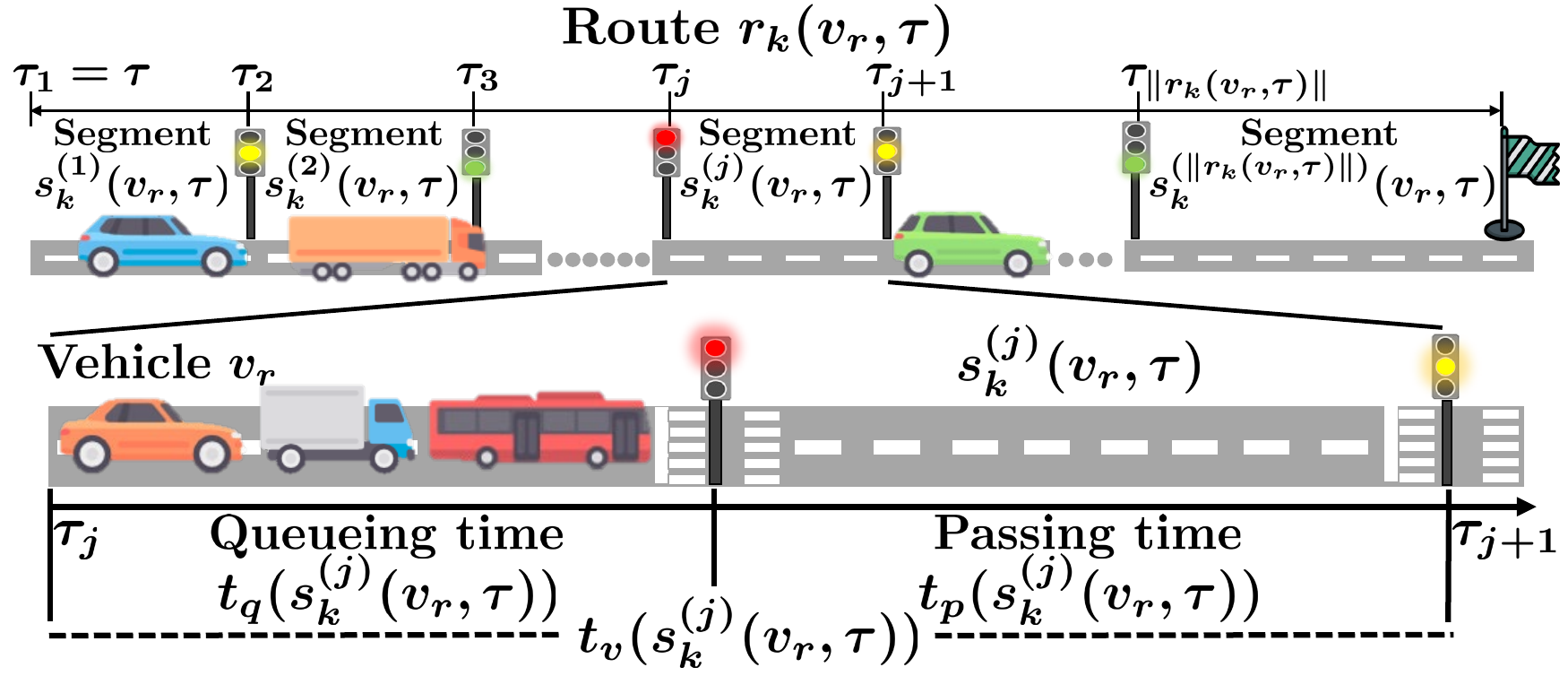}
\caption{Breakdown of segment travel‑time components: queueing delay $t_q$ and passing time $t_p$.}
\label{fig:Trv_Time_Segs}
\end{figure}
\FloatBarrier

Let $\tau_j$ represent the time instant when vehicle $v_r$ arrives before entering segment $s_k^{(j)}(v_r,\tau)$. After queue clearance time $\tau_{(j)}=\tau_j+t_q\bigl(s_k^{(j)}(v_r,\tau)\bigr)$, the effective forecast horizon index is:
\begin{equation}\label{eq: hj_ttp}
h_j=\min\left(T_{\mathrm{out}}-1,\left\lfloor\frac{\tau_{(j)}-\tau}{T_h}\right\rfloor_{+}\right),
\end{equation}
where $[\cdot]_{+}$ clamps negative values to zero. The segment passing‑time is calculated as:
\begin{equation}\label{eq: tp_hybrid}
t_p\bigl(s_k^{(j)}(v_r,\tau)\bigr)
=\frac{L_{rs}\bigl(s_k^{(j)}(v_r,\tau)\bigr)}{\max\bigl(\tilde{V}\bigl(s_k^{(j)}(v_r,\tau),\tau,h_j,v_r\bigr),\,\varepsilon_{\mathrm{spd}}\bigr)}.
\end{equation}
Here $\varepsilon_{\mathrm{spd}}$ denotes a small positive speed floor ensuring numerical stability.

\subsection{Thin‑Boundary Cost‑Announcement Interface Model}
\label{sec:interface}
This subsection formalizes the exclusive cross‑partition communication specification enforced over the thin boundary, as visualized in Fig.~\ref{fig:cosim-panels}.

Let $\mathcal{V}$ denote the full set of civilian vehicles within the intact network, and $\mathcal{V}_{\mathrm{free}}\subseteq\mathcal{V}$ denote uncommitted civilian vehicles that have not yet been assigned to carry DTN bundles. Define $\pi_{\mathrm{via}}(v;g,u)$ as the detour route for vehicle $v$ that traverses gateway $g$ and remote gateway $u$, then continues toward $v$'s original destination. The hybrid cost metric $t_{\mathrm{hyb}}(\pi)$ combines live and forecast speeds, excluding both driver‑aware scaling factor $\kappa$ and queueing delay:
\begin{multline}
t_{\mathrm{hyb}}(\pi)
=\sum_{s\in\pi} L_{rs}(s) \big/ \max\bigl(\alpha(h_s)\,\bar{V}_{\mathrm{o}}(s,\tau) \\
+(1-\alpha(h_s))\,\bar{V}_{\mathrm{f}}(s,\tau,h_s),\,\varepsilon_{\mathrm{spd}}\bigr).
\label{eq:blend}
\end{multline}

The detour overhead $\Delta(v;g,u)$ measures the extra travel cost when vehicle $v$ takes the two‑gateway detour $\pi_{\mathrm{via}}(v;g,u)$ instead of its original residual route:
\begin{equation}
\Delta(v;g,u)=t_{\mathrm{hyb}}\big(\pi_{\mathrm{via}}(v;g,u)\big)-t_{\mathrm{hyb}}\big(r(v,\tau)\big).
\label{eq:extra}
\end{equation}

The scalar cost‑announcement broadcast by gateway $g$ targeting remote gateway $u\in\mathcal{U}_q$ is the minimal detour overhead among all uncommitted civilian vehicles:
\begin{equation}
c_{g,u}=\min_{v\in\mathcal{V}_{\mathrm{free}}}\Delta(v;g,u).
\label{eq:quote}
\end{equation}
The DTN‑UAV Forwarder consumes this scalar value directly. The beacon $\mathcal{B}_g(t)$ aggregates all valid $(u,c_{g,u})$ pairs.

We define $\mathcal{E}_{\mathrm{gw}}(g,u)$ as the \emph{gateway‑arrival event}: a signal sent from a damaged corner to notify the intact side that a bundle has been buffered at gateway $g$ and is destined for remote gateway $u$. Only two message types are permitted to cross the thin boundary (depicted in Fig.~\ref{fig:cosim-panels}):
\begin{enumerate}
\item Intact side $\rightarrow$ corner: beacon entry $(u,c_{g,u})$, containing target‑gateway index and scalar cost‑announcement.
\item Corner $\rightarrow$ intact side: gateway‑arrival event $\mathcal{E}_{\mathrm{gw}}(g,u)$.
\end{enumerate}

These message definitions constitute the formal cross‑partition interface between intact‑network modules and corner‑local modules.

\subsection{DTN‑UAV Forwarder}
\label{sec:dtn}
\label{sec:corner}

The DTN‑UAV Forwarder executes inside each damaged corner (see Fig.~\ref{fig:cosim-panels}). It acts as a fixed simulation testbed component rather than a newly designed DTN routing algorithm; its internal logic remains identical across all comparative experiments. Selective‑Rerouter computes minimal two‑gateway detour costs within the intact network, while the forwarder passively ingests published cost‑announcement beacons without recomputing the minimization in \eqref{eq:quote}.

Inter‑node communication inside a corner occurs only when pairwise node distance falls below radio range $R$. Physically separated damaged corners have no direct wireless links. UAVs are confined within their host corner and its adjacent boundary corridor; neither UAVs nor ground‑patrol oscillators can enter intact road segments (illustrated in Fig.~\ref{fig:cosim-panels}).

Every $T_{\mathrm{ref}}$ simulation steps, gateway $g$ re‑evaluates $c_{g,u}(t)$ following~\eqref{eq:quote} and publishes the updated beacon:
\begin{equation}
\mathcal{B}_{g}(t)
=
\bigl\{
(u,\,c_{g,u}(t))
:\ u\in\mathcal{U}_{q},\ \ \mathcal{V}_{\mathrm{free}}(t)\neq\emptyset
\bigr\}.
\label{eq:beacon}
\end{equation}
Here $c_{g,u}(t)$ represents the minimal detour overhead from gateway $g$ toward remote gateway $u$. If no feasible uncommitted civilian vehicle exists for a target $u$, that entry will not appear in the beacon.

Corner‑local nodes maintain local caches for incoming cost‑announcements. For node $i$, the cache is initialized empty, and unknown cost values are treated as $+\infty$:
\begin{equation}
\mathcal{C}_{i}(t)=\{(g,u,\bar{c}_{i,(g,u)}(t))\}.
\label{eq:cache}
\end{equation}
Upon receiving a new announcement $(u,c_{g,u})$ from gateway $g$: the triple is inserted for the first arrival; otherwise the cached value is updated via
\begin{equation}
\bar{c}_{i,(g,u)}(t)
\leftarrow
\min\{\bar{c}_{i,(g,u)}(t),\,c_{g,u}\}.
\label{eq:cacheupd}
\end{equation}
Cached cost values evolve in a monotonically non‑increasing fashion. Equal or larger incoming cost values will not overwrite existing cache entries. When holding bundle $m$, node $i$ selects the exit gateway according to cached cost entries:
\begin{equation}
g^{\dagger}(i,m)
=
\operatorname*{arg\,min}_{\substack{g\in\mathcal{G}_{q}\\(g,u_{m})\in\mathcal{C}_{i}(t)}}
\bar{c}_{i,(g,u_{m})}(t),
\label{eq:exit}
\end{equation}
Gateways without cached records for target $u_m$ are excluded from candidates.

Each bundle carries structured metadata:
\begin{equation}
\beta_{m}=(\mathrm{id}_{m},\, b_{m},\, o_{m},\, u_{m},\, o^{\star}_{m}),
\label{eq:bundle}
\end{equation}
where $\mathrm{id}_{m}$ is bundle identifier, $b_{m}$ is auxiliary payload, $o_{m}$ is the source ground‑patrol oscillator generating bundle $m$, $u_{m}$ is target gateway index, and $o^{\star}_{m}$ denotes the destination ground‑patrol oscillator that finally consumes the bundle. Gateways serve purely as cross‑corner transfer points and are not final destinations.

The source oscillator $o_m$ and target gateway $u_m$ are fixed at bundle generation. In contrast, the exit gateway is dynamically chosen by formula~\eqref{eq:exit}. The intact‑side module learns this exit gateway when receiving the gateway‑arrival event $\mathcal{E}_{\mathrm{gw}}(g,u_m)$.

Intra‑corner message delivery adopts store‑carry‑forward mechanisms. Each UAV maintains separate inbound and outbound buffers, and scheduling prioritizes bundles with shorter remaining lifetime. Once a bundle arrives inside a gateway buffer, the DTN‑UAV Forwarder emits event $\mathcal{E}_{\mathrm{gw}}(g,u)$ across the thin boundary.

As depicted in Fig.~\ref{fig:cosim-panels}, each damaged corner is equipped with one UAV $a_q$ and a set of ground‑patrol oscillators $\mathcal{O}_q$. Ground‑patrol oscillators generate and receive DTN bundles along fixed trajectories, yet direct radio contact with gateways is not always guaranteed due to limited communication coverage. Benefiting from high mobility, the UAV acts as an intra‑corner relay: it collects bundles from ground oscillators and delivers them to gateways upon radio contact, and forwards bundles arriving from gateways toward destination oscillator $o_{m}^{\star}$.

\section{Problem Formulation}
\label{sec:problem}
This section defines the cross‑corner routing problem built upon the system model of Section~\ref{sec:model}.
A key distinguishing feature of the studied post‑disaster setting is the mixed‑mode delivery paradigm: DTN bundles generated within damaged subregions are ferried across partition boundaries by civilian vehicles travelling over the intact road network.
This section formalizes the optimization objectives, practical challenges arising in partitioned urban environments, and feasibility constraints that any valid rerouting policy must satisfy under the thin‑boundary interface from Section~\ref{sec:interface}. It also clarifies the problem scope and underlying assumptions.
Section~\ref{sec:selch} presents one feasible rerouting policy satisfying all constraints.
The evaluation protocol, performance metrics, and baseline‑algorithm eligibility criteria for comparative experiments are given in Section~\ref{sec:paradigm}.

\subsection{Cross‑Corner Optimization Objective and Problem Challenges}
\label{sec:objective}
\label{sec:challenges}
Each bundle $m$ belonging to the full message set $\mathcal{M}$ originates at a source ground‑patrol oscillator $o_{m}$, and its final destination is a fixed ground‑patrol oscillator $o^{\star}_m$ located inside a different damaged corner.
As a hard input constraint of the problem formulation: \textit{bundle traffic cannot target gateways belonging to its source corner}. Therefore, a bundle can only be addressed to a gateway drawn from the remote‑gateway set $\mathcal{U}_q$ defined in \eqref{eq:destset}, whose elements correspond exclusively to gateways residing in non‑source corners, together with the target ground‑patrol oscillator inside that remote corner.
Bundles are uniformly distributed among all remote non‑source corners, and intra‑corner bundle transmission is disabled.

The Selective‑Rerouter module has read‑access to vehicle trajectories, road‑segment occupancy, and speed measurements collected on the intact‑network side, and it is capable of computing and assigning travel routes for civilian vehicles on intact roads.
However, it has no visibility regarding which internal node inside a damaged corner currently holds an undelivered bundle.
Every bundle $m$ must arrive at its destination oscillator $o^{\star}_m$ no later than its time‑to‑live $T_{\mathrm{TTL}}$.

The entity under comparative evaluation is the intact‑side rerouting policy. For a given set of bundles $\mathcal{M}$, the multi‑objective optimization targets are formulated as:
\begin{align}
\min \quad & \bar{T}_{\mathrm{obj}},\; T_{\mathrm{obj}}^{\max},\; \bar{t}_{\mathrm{civ}}, \label{eq:objmin}\\
\max \quad & D_{\mathrm{osc}},\; D_{\mathrm{osc}}/M, \label{eq:objmax}
\end{align}

These optimization objectives are mutually conflicting. Shortening bundle end‑to‑end delivery latency may force civilian vehicles to take longer detours over intact road segments, which consequently increases the average civilian trip duration $\bar{t}_{\mathrm{civ}}$.
Algorithm performance is characterized by the objective vector formed by \eqref{eq:objmin}--\eqref{eq:objmax}.
Any candidate rerouting policy must comply with constraints \eqref{eq:cbeacon}--\eqref{eq:cfree}, the bundle lifetime bound $T_{\mathrm{TTL}}$, and the deferred‑assignment rule: a dedicated two‑gateway civilian route for bundle $m$ shall be generated \textit{only after} bundle $m$ has arrived at a gateway buffer.

Here $M=|\mathcal{M}|$ denotes the total count of bundles generated in the system.
$D_{\mathrm{osc}}$ counts the number of bundles whose copy successfully reaches the destination ground‑patrol oscillator $o^{\star}_m$;
$D_{\mathrm{osc}}/M$ represents the overall bundle delivery ratio.
$\bar{T}_{\mathrm{obj}}$ and $T_{\mathrm{obj}}^{\max}$ are respectively the mean and maximum birth‑to‑reception latency computed over all successfully delivered bundles.
$\bar{t}_{\mathrm{civ}}$ is the average travel duration of completed civilian vehicle trips.
Detailed counting rules for these metrics are provided in Section~\ref{sec:metrics}.

A target gateway is fixed upon bundle generation. A gateway is eligible for selection only if civilian vehicles can exit the gateway onto intact road edges, and the gateway is reachable from every other damaged corner via the intact‑network topology.
All damaged road edges are excluded both from cost‑announcement computation and from all vehicle routes output by the rerouting policy.

In addition to the above multi‑objective formulation, this hypothetical partitioned post‑disaster scenario inherits well‑known practical shortcomings from existing rerouting solutions designed for fully connected, undamaged cities. Additional difficulties are introduced jointly by thin‑boundary information‑isolation restrictions and the mixed delivery workflow where civilian mobility carries DTN bundles across urban partitions.
Key practical challenges are summarized below:
\begin{itemize}
    \item Relying purely on road‑segment occupancy fails to detect slow‑moving queues that have not yet fully filled the physical road segment.
    \item Static fixed‑size upstream intervention windows select inappropriate sets of candidate vehicles for rerouting.
    \item Path‑cost evaluation relying solely on instantaneous live speed, or forecasts trained only for segment‑level regression error, produces incorrect relative ordering among candidate travel paths.
    \item Adopting a uniform speed model for all drivers ignores the natural variance observed in real‑world individual driving trajectories.
    \item Performing replanning for nearly all vehicles wastes computational resources on trips that will never encounter traffic bottlenecks.
\end{itemize}
Prior intact‑network rerouting works surveyed in Section~\ref{sec:related} leave these practical challenges unresolved.
The next subsection defines the mandatory constraints that all admissible rerouting policies must satisfy.

\subsection{Thin‑Boundary Feasibility Constraints}
\label{sec:boundary}
Building upon the thin‑boundary cost‑announcement interface defined in Section~\ref{sec:interface}, we state the feasibility constraints that every admissible rerouting policy must obey.
A policy that violates any constraint below lies outside the solution space of this problem.

Deferred binding forbids assigning a civilian carrier or constructing a two‑gateway vehicle route before the gateway‑arrival event $\mathcal{E}_{\mathrm{gw}}(g,u)$ defined in Section~\ref{sec:interface}.
The corner‑local cache update rule in \eqref{eq:cacheupd} accepts an incoming cost‑announcement if and only if the received scalar cost value is strictly smaller than the cached entry. As a result, the selected exit gateway $g^{\dagger}(i,m)$ will not be altered upon receiving cost‑announcements of equal or larger numerical value.

Only two types of messages are permitted to traverse the thin boundary; internal state information belonging to either partition cannot leak across this boundary.
Full message syntax specifications are given in Section~\ref{sec:interface}.
A feasible policy must satisfy the following set of formal constraints.

The beacon constraint allows gateway $g$ to publish only
\begin{equation}
\mathcal{B}_g(t)\subseteq\big\{(u,c_{g,u}):u\in\mathcal{U}_q\big\},
\label{eq:cbeacon}
\end{equation}
where $c_{g,u}$ is defined in \eqref{eq:quote}; the gateway transmits no beacon entry for target $u$ when no valid solution exists.
The beacon payload shall carry no civilian‑vehicle identifiers, route descriptions, nor geographic coordinates of damaged corners.

The arrival constraint allows a damaged corner to report only
\begin{equation}
(g,u),
\label{eq:carrival}
\end{equation}
which signals the event: a copy of the bundle has entered the buffer of gateway $g$, destined for remote gateway $u$.
This cross‑boundary notification shall not expose internal identifiers of nodes within damaged corners.
Probe messages carrying route information stay entirely within the intact network and contain no bundle payload field.

The binding constraint enforces the deferred‑binding principle described above.
For any bundle $m$, before the gateway‑arrival event occurs, no vehicle from the free‑civilian set $\mathcal{V}_{\mathrm{free}}$ shall be pre‑committed to serve as its carrier.
After receiving the arrival report $(g,u)$, Selective‑Rerouter may generate one valid both‑gateway route for a qualified civilian vehicle $v$, and it must perform the set update:
\begin{equation}
\mathcal{V}_{\mathrm{free}}\leftarrow\mathcal{V}_{\mathrm{free}}\setminus\{v\}.
\label{eq:cfree}
\end{equation}

Subsequent cost‑announcement computations operate over this updated, reduced set of free civilian vehicles.
Congestion‑triggered ordinary replanning may modify routes for uncommitted free civilians as well as vehicles that have acquired bundles. Only vehicles that have already been formally assigned as dedicated bundle carriers are excluded from this replanning procedure.
This general congestion‑driven replanning operation is distinct from the dedicated both‑gateway route generation for bundle transportation.
Any rerouting scheme assigning civilian carriers in advance of bundle gateway‑buffer arrival is ineligible for end‑to‑end comparative evaluation.

\subsection{Problem Scope and Assumptions}
\label{sec:scope}

\subsubsection{Assumptions of This Statement}
\label{sec:publishers}
Every simulation run reported in this paper adopts the cost‑announcement definition given by \eqref{eq:quote} from Section~\ref{sec:interface}.
Gateway $g$ computes the minimal detour overhead over the set $\mathcal{V}_{\mathrm{free}}$ and publishes the corresponding beacon, or remains silent for target $u$ if no feasible free civilian vehicle exists.
Published beacons transmit only scalar cost values and gateway indices.
The hybrid travel‑time metric $t_{\mathrm{hyb}}$ used inside detour cost $\Delta(v;g,u)$ follows definition \eqref{eq:blend}.
Damaged road segments are excluded from all relevant calculations.
Assignment of a civilian carrier occurs strictly after receiving gateway‑arrival event $(g,u)$.
The detailed definition for cost‑announcements is not repeated here.

Within this problem formulation, the DTN‑UAV Forwarder inside each damaged corner is treated as a fixed black‑box component; optimization is applied exclusively to the rerouting policy executing on the intact‑network side.

\subsubsection{Outside This Statement}
Performance statistics collected only from within damaged corners shall not be counted as end‑to‑end evaluation metrics described in Section~\ref{sec:metrics}.
Simulation scenarios without damaged corners evaluate only civilian trip durations, since no cross‑corner DTN bundles are injected.
The following settings lie outside the scope of the present formulation: alternative partitioned road‑network topologies, cost‑announcement rules dynamically varying with civilian fleet conditions, and wireless contact models with packet loss.
Potential future extensions covering these directions are noted in Section~\ref{sec:conclusion}.

\section{Selective‑Rerouter}
\label{sec:selch}
This section presents Selective‑Rerouter, the proposed rerouting policy solving the multi‑objective optimization problem formulated in Section~\ref{sec:objective}, while complying with all thin‑boundary feasibility constraints from Section~\ref{sec:boundary}. Each core component addresses challenges in Section~\ref{sec:challenges}: dual‑metric congestion detection, severity‑adaptive upstream intervention window, horizon‑aware hybrid live–forecast path costing with path‑ranking loss, plus driver‑aware forecast scaling preserving raw live traffic observations. In benchmark comparisons, the DTN‑UAV Forwarder (Section~\ref{sec:dtn}) remains fixed. Only Selective‑Rerouter is substituted for evaluation. Algorithm~\ref{alg:reroute} defines the top‑level execution step, and Algorithm~\ref{alg:iface} implements the \emph{thin‑boundary} beacon publication and $\mathcal{V}_{\mathrm{free}}$ maintenance logic. Inner‑stage vehicle‑selection and route‑allocation logic are described textually in Sections~\ref{subsec:rvs} and~\ref{subsec:phase3_ara}.

Congestion is detected jointly from occupancy and speed measurements. The upstream intervention window expands proportionally to congestion severity, incorporating vehicles approaching bottlenecks within $\theta_{\mathrm{app}}$ hops. Path evaluation fuses real‑time traffic observations and short‑horizon forecasts, where blending weight $\alpha(h)$ decays with prediction horizon. Only the prediction head of LSTAN‑GERPE~\cite{WangYang2025LSTAN} receives path‑ranking‑loss fine‑tuning. The driver‑aware scaling factor modifies solely forecast outputs without altering live measurements. The algorithm further computes \emph{cost-announcement} values and constructs both‑gateway routes upon incoming \emph{gateway}‑arrival events.

\begin{algorithm}[!t]
\LinesNumbered
\KwIn{intact fleet $\mathcal{V}$, arrival events from this step}
\KwOut{costs $c_{g,u}$, written routes, and named civilians}
Detect congestion, select vehicles, and allocate routes following Section~\ref{pha:tcd}, Section~\ref{subsec:rvs}, and Section~\ref{subsec:phase3_ara}\;
\ForEach{gateway $g$ and destination $u$}{
  Set $c_{g,u}$ by \eqref{eq:quote}, or leave it undefined\;
  Retain the argmin identity only inside this procedure\;
}
\ForEach{arrival $(g,u)$ whose bundle still has no written route}{
  Write one both‑gateway route by Section~\ref{sec:intact}\;
  Return the named civilian to the interface\;
}
\caption{Selective‑Rerouter: one step. A literature policy replaces this procedure and leaves the DTN‑UAV Forwarder unchanged.}
\label{alg:reroute}
\end{algorithm}

\begin{algorithm}[!t]
\LinesNumbered
\KwIn{costs and named civilians returned by \textsc{Reroute}}
\KwOut{beacon $\mathcal{B}_g(t)$, updated $\mathcal{V}_{\mathrm{free}}$}
\If{$t \bmod T_{\mathrm{ref}} = 0$}{
  Re-publish each defined $c_{g,u}$ on the beacon of gateway $g$\;
  Discard every argmin vehicle before the beacon is formed\;
}
Remove each civilian named for an arrival from $\mathcal{V}_{\mathrm{free}}$\;
\caption{Interface handler. The DTN‑UAV Forwarder receives the beacon and does not receive the named civilian.}
\label{alg:iface}
\end{algorithm}

\subsection{Path‑Cost Evaluation}
\label{subsec:forecast_cost}
\label{subsec:pred_pipeline}
Graph definitions, occupancy‑speed metrics, hybrid speed formulation and segment travel‑time relations are inherited from Section~\ref{sec:system_model}. Algorithm~\ref{alg:hybrid_pathsum} computes the passing‑time score $C_r^{p}(r_k)$ for candidate route $r_k$, with horizon‑dependent blending weight $\alpha(h)=\max(0,\alpha_0-h\cdot\alpha_{\mathrm{dec}})$, corresponding to the passing‑time term in \eqref{eq: crt}.

\begin{algorithm}[!t]
\LinesNumbered
\KwIn{$f_{\theta}$, $\mathbf{X}_{\tau}$, $\bar{V}_{\mathrm{o}}$, candidate $r_k$, $Bd$, $\kappa$, $(\alpha_0,\alpha_{\mathrm{dec}})$}
\KwOut{$C_r^{p}(r_k)$}
\Begin{
    $\bar{V}_{\mathrm{f}}\gets f_{\theta}(\mathbf{X}_{\tau};L_{\mathrm{lap}})$\tcp*{batched multi‑horizon fleet‑mean forecast}
    $C_r^{p}\gets 0$\; $\tau_1\gets\tau$\tcp*{$\tau_1$: arrival at first‑segment intersection}
    \For{$j\gets 1$ \KwTo $\lVert r_k\rVert$}{
        Compute $t_q\bigl(s_k^{(j)}\bigr)$ via the traffic‑light model of~\cite{8641378}\;
        Set $h_j$ by~\eqref{eq: hj_ttp}\;
        $\alpha(h_j)\gets\max\bigl(0,\alpha_0-h_j\cdot\alpha_{\mathrm{dec}}\bigr)$\tcp*{live‑measurement weight}
        Form $\hat{V}_{\mathrm{f}}\bigl(s_k^{(j)},\tau,h_j,v_r\bigr)$ by~\eqref{eq:hatV_def}\;
        Form $\tilde{V}\bigl(s_k^{(j)},\tau,h_j,v_r\bigr)$ by~\eqref{eq: hybrid_speed}\;
        $t_p\gets L_{rs}\bigl(s_k^{(j)}\bigr)/\max(\tilde{V},\varepsilon_{\mathrm{spd}})$\tcp*{passing time~\eqref{eq: tp_hybrid}}
        $C_r^{p}\gets C_r^{p}+t_p$\;
        $\tau_{j+1}\gets\tau_j+t_q\bigl(s_k^{(j)}\bigr)+t_p$\;
    }
    \Return{$C_r^{p}(r_k)$}\;
}
\caption{Hybrid path sum $C_r^{p}(r_k)$ (Phase~3 passing‑time score with $\tilde{V}$)}
\label{alg:hybrid_pathsum}
\label{alg:TimeCost}
\end{algorithm}

\subsection{Forecast and Driver Scale}
\label{sec:travel_time_prediction}
This subsection defines forecast output $\bar{V}_{\mathrm{f}}$ and driver‑aware scaling coefficient $\kappa$ used within \eqref{eq:hatV_def}. Pure segment‑level regression loss may invert relative travel‑time ordering for full candidate routes. Hence our pipeline combines traffic forecaster output with per‑driver scaling adjustments.

\subsubsection{Network‑Level Speed Forecasting}
\label{subsec:lstan_gerpe}
\paragraph{Input–output formulation.}
Let $\mathcal{S}$ denote road‑segment set defined in Sec.~\ref{sec:system_model}, $N=|\mathcal{S}|$. At decision instant $\tau$, loop‑detector observations construct input tensor:
\begin{equation}\label{eq:forecaster_input}
\mathbf{X}_{\tau}=
\bigl\{x(s,\tau-iT_h)\bigm|\, s\in\mathcal{S},\;
i=0,\dots,T_{\mathrm{in}}-1
\bigr\}
\in\mathbb{R}^{T_{\mathrm{in}}\times N\times F},
\end{equation}
where $x(s,\tau-iT_h)$ stacks $\bar{V}_{\mathrm{o}}$ and $R_O$, $T_{\mathrm{in}}$ denotes history length in forecast steps $T_h$, and $F{=}2$. Lag $i{=}0$ marks observation window ending at $\tau$. We adopt \texttt{LSTAN\_GERPE}~\cite{WangYang2025LSTAN} equipped with Laplacian graph encoding $L_{\mathrm{lap}}$:
\begin{equation}\label{eq:forecaster_output}
\bar{V}_{\mathrm{f}}(s,\tau,h)=f_{\theta}\bigl(\mathbf{X}_{\tau};\,L_{\mathrm{lap}}\bigr)_{s,h},\qquad
h=0,\dots,T_{\mathrm{out}}-1,
\end{equation}
producing fleet‑mean predicted speed for every segment and forecast horizon. Original Huber pre‑training, network architecture and hyper‑parameters follow the published reference~\cite{WangYang2025LSTAN}. Only the prediction head undergoes downstream fine‑tuning.

\paragraph{Reroute‑aligned ranking fine‑tuning.}
Stage‑1 optimizes $f_{\theta}$ using only Huber speed loss ($\lambda_{\mathrm{rank}}{=}0$). Stage‑2 freezes encoder weights and updates exclusively the prediction head under reduced learning rate:
\begin{equation}\label{eq:rank_loss}
\mathcal{L}=\mathcal{L}_{\mathrm{Huber}}+\lambda_{\mathrm{rank}}\,\mathcal{L}_{od\_ksp},
\end{equation}
with small positive $\lambda_{\mathrm{rank}}$. For each historical OD pair $\omega$, pre‑compute Yen $k$‑shortest‑path set $\mathcal{P}_{\omega}$~\cite{10.2307/2629312} matching online candidate‑path budget $N_{cr}$. The ranking surrogate omits queueing term $t_q$, speed blending and driver scaling:
\begin{equation}\label{eq:path_travel_time}
t_{\mathrm{len}}(P;V)=\sum_{s\in P}\frac{L_{rs}(s)}{\max\bigl(V(s),\varepsilon_{\mathrm{spd}}\bigr)},
\end{equation}
$\varepsilon_{\mathrm{spd}}=10^{-4}\,\mathrm{m/s}$ (Table~\ref{tab:params}). Predicted and ground‑truth path durations:
\begin{equation}\label{eq:path_time_pred}
\hat{t}_{\mathrm{len}}(P;\tau,h)=t_{\mathrm{len}}\bigl(P;\bar{V}_{\mathrm{f}}(\cdot,\tau,h)\bigr),
\end{equation}
\begin{equation}\label{eq:path_time_gt}
t_{\mathrm{len}}^{*}(P;\tau)=t_{\mathrm{len}}\bigl(P;\bar{V}_{\mathrm{o}}(\cdot,\tau)\bigr).
\end{equation}
Path pairs satisfying minimum time gap threshold $t_{\min}$ contribute pairwise loss:
\begin{equation}\label{eq:od_ksp_pair}
\begin{aligned}
\mathcal{L}_{\mathrm{pair}}(P,P';\tau,h)
&=\log\Bigl(1+\exp\bigl(-\mathrm{sign}\bigl(\Delta t^{*}_{P,P'}(\tau)\bigr)\\
&\qquad\cdot\Delta\hat{t}_{P,P'}(\tau,h)\bigr)\Bigr),
\end{aligned}
\end{equation}
where $\Delta t^{*}_{P,P'}$ and $\Delta\hat{t}_{P,P'}$ denote ground‑truth and predicted path‑time gaps. Aggregate across forecast horizons, OD mini‑batches and valid path pairs:
\begin{equation}\label{eq:od_ksp_loss}
\mathcal{L}_{\mathrm{od\_ksp}}
=\frac{1}{|\mathcal{H}|\,|\mathcal{W}|}\sum_{h\in\mathcal{H}}\sum_{\omega\in\mathcal{W}}
\frac{1}{|\mathcal{A}_{\omega}|}\sum_{(P,P')\in\mathcal{A}_{\omega}}\mathcal{L}_{\mathrm{pair}}(P,P';\tau,h),
\end{equation}
with $\mathcal{H}=\{0,\dots,T_{\mathrm{out}}-1\}$.

\subsubsection{Driver‑Behavior Personalization}
\label{subsec:dtm}
Driver mapping parameter $Bd$ and scaling factor $\kappa$ are derived from historical per‑vehicle traversal records. Fallback values apply for insufficient history: $Bd(v_r,s)\gets 1/2$ for available segment statistics. $\kappa(v_r,s)\gets 1$ when vehicle‑level samples are unreliable (Proposed-noDB mode). Raw segment traversal speed:
\begin{equation}\label{eq:6.1}
V_{\mathrm{hist}}(s,\tau, v_r) = \frac{L_{rs}(s)}{t_{r}^{v_r}(\tau_x) - t_{r}^{v_r}(\tau_y)},
\end{equation}
$t_r^{v_r}(\tau_x)$, $t_r^{v_r}(\tau_y)$ are segment exit and entry timestamps. Historical average $V_{\mathrm{hist}}^{\mathrm{avg}}(s,v_r)$ differs from instantaneous observed fleet speed $\bar{V}_{\mathrm{o}}$. Normalized driver index:
\begin{equation}\label{eq:Bdcalculation}
Bd(v_r,s)= \frac{V_{\mathrm{hist}}^{\mathrm{avg}}(s, v_r) - V_{\mathrm{hist}}^{\mathrm{avg}}(s, v_{\mathrm{lo}})}{V_{\mathrm{hist}}^{\mathrm{avg}}(s, v_{\mathrm{hi}}) - V_{\mathrm{hist}}^{\mathrm{avg}}(s, v_{\mathrm{hi}})}.
\end{equation}
Per‑segment ratio samples $z_j^{s}=V_{\mathrm{hist}}/\bar{V}_{\mathrm{o}}$ undergo normalization:
\begin{equation}\label{eq:62}
\begin{aligned}
z_{j}^{*s} = &\frac{z_{j}^{s} - z_{\min}^{s}}{z_{\max}^{s} - z_{\min}^{s}} \times \frac{n_{s} - 1}{n_{s}} + \frac{1}{2 \times n_{s}},\\
z_{\max}^{s} = &\max\limits_{j=0,\dots,n_{s} - 1} z_{j}^{s},\\
z_{\min}^{s} = &\min\limits_{j=0,\dots,n_{s} - 1} z_{j}^{s}.
\end{aligned}
\end{equation}
Empirical ratio distribution follows power‑law statistics~\cite{akaike1998information}:
\begin{equation}\label{eq:eta}
\eta_s = -\frac{n_{s}}{\sum_{j=0}^{n_{s}-1}\ln z_{j}^{*s}}.
\end{equation}
Calibrated per‑driver scaling coefficient:
\begin{equation}\label{eq:delta}
\kappa(v_r,s) = \big[(z_{\max}^{s} - z_{\min}^{s}) \times (Bd(v_r,s))^{\eta_s} + z_{\min}^{s}\big].
\end{equation}
Factor $\kappa$ exclusively rescales forecast outputs in \eqref{eq:hatV_def}, without modifying live observations.

\subsection{Cost-Announcement, Selection, and Deferred Write}
\label{sec:reroute}
The rerouting pipeline runs at the decision period $T=300\,\mathrm{s}$ (Table~\ref{tab:params}), not on every simulation step. The DTN‑UAV Forwarder consumes only the scalar cost-announcement $c_{g,u}$ and post‑arrival edge‑sequence routes, and has no access to intermediate rerouting state.

\paragraph{Detection and selection.}
Congestion detection and severity scoring follow Section~\ref{pha:tcd}, including soft‑threshold triggering rules. Vehicle‑selection and allocation hyper‑parameters are summarized in Table~\ref{tab:params}. At $\tau+T_{\mathrm{alert}}$, if bottleneck conditions persist on original segments, candidate pools are re‑evaluated using forecasts computed at time $\tau$.

\subsubsection{Cost-Announcement and Deferred Write}
\label{sec:intact}
A \emph{gateway}‑arrival event triggers civilian‑carrier assignment. Ordinary congestion‑driven replanning is an independent decision. The \emph{corner} side observes neither replanning logic, only published $c_{g,u}$ before bundle hand‑off.

\paragraph{Both‑gateway path.}
On dual graph $G$ (Section~\ref{sec:system_model}), a valid both‑gateway route for civilian $v$, source gateway $g$, and remote gateway $u$ concatenates three Dijkstra shortest‑path segments under free‑flow length $L_{rs}$: from the vehicle's current position to gateway $g$, from gateway $g$ to gateway $u$, and from gateway $u$ to the civilian's original destination. Incomplete path components disqualify $v$ as a cost-announcement candidate, complying with constraints in Section~\ref{sec:boundary}. This path construction is reused both during cost-announcement computation and after gateway-arrival assignment. The civilian minimizer identity is discarded for beacon publication and only retained internally for post‑arrival route generation.

\paragraph{Cost-announcement travel time.}
The published cost-announcement does not use ranked candidate‑path costs. It evaluates $t_{\mathrm{hyb}}$ defined in \eqref{eq:blend}. Minimization runs over the full $\mathcal{V}_{\mathrm{free}}$ set of uncommitted civilians (not limited to reroute‑selected vehicle pool). Gateways emit no beacon entry when no feasible free civilian exists.

The assigned‑carrier route cost $t_{\mathrm{asgn}}(v)$ includes signal queueing delay and driver‑aware scaling via blended speed $\tilde{V}$:
\begin{equation}
t_{\mathrm{asgn}}(v)
=\sum_{j}\left(
t_q\bigl(s^{(j)}\bigr)
+\frac{L_{rs}\bigl(s^{(j)}\bigr)}{\max\bigl(\tilde{V}(s^{(j)},\tau,h_j,v),\,\varepsilon_{\mathrm{spd}}\bigr)}
\right),
\label{eq:asgn}
\end{equation}
Phase‑3 candidate ranking uses $C_r^{p}$ which excludes queueing $t_q$. Cost-announcement computation $t_{\mathrm{hyb}}$ excludes both $t_q$ and $\kappa$ scaling. Subsequent decision instants may recompute segments subject to which node currently holds the bundle.

\paragraph{One‑shot carrier after handover.}
Let $t_{\mathrm{gw}}(m)$ denote the time at which bundle $m$ enters the buffer of gateway $g$ for destination gateway $u$. No civilian may be assigned for $t<t_{\mathrm{gw}}(m)$. Upon arrival, free civilians are sorted by spatial gap between their remaining route and destination‑gateway coordinates:
\begin{equation}
\gamma(v)=\min_{s\in r(v,t_{\mathrm{gw}}(m))}\lVert \mathrm{cent}(s)-y^{\star}\rVert_2,
\label{eq:gap}
\end{equation}
where $\mathrm{cent}(s)$ is segment centroid and $y^{\star}$ destination‑gateway coordinate. We test first $N_{\mathrm{try}}$ civilians ordered by ascending $\gamma(v)$. The first civilian satisfying both‑gateway path constraints is selected and removed from $\mathcal{V}_{\mathrm{free}}$.

\paragraph{Four vehicle states.}
Benchmark rerouting policies maintain separate scoring logic for uncommitted civilians, civilians having completed bundle delivery, vehicles en‑route toward source gateway, and civilians actively carrying bundles. After bundle delivery at destination gateway, the vehicle reverts to uncommitted‑civilian processing rules.

\paragraph{Named oscillator after the destination gateway.}
Source oscillator $o_{m}$, target gateway $u_{m}$, and destination oscillator $o^{\star}_m$ are fixed at bundle generation. Bundle delivery completes when oscillator $o^{\star}_m$ lies within radio‑range $R$ of destination gateway:
\begin{equation}
\lVert p_{o^{\star}_m}-p_{u_m}\rVert_2\le R.
\label{eq:named}
\end{equation}
Otherwise, destination‑corner UAV forwards the oldest waiting bundle exclusively to $o^{\star}_m$.
Cost-announcement generation follows cross‑boundary rules. An alternative rank‑mode publisher may broadcast gateway rank instead of raw time values while retaining gateway source identifier.

\subsection{Three‑Phase Pipeline}
\label{sec_alg}
\textsc{Reroute} triggers replanning only for vehicles affected by detected congestion. At decision instant $\tau$, it produces congested‑segment set $\mathcal{C}(\tau)$, candidate vehicle pool $S_{rv}^{\mathrm{tot}}(\tau)$ and updated output routes. Fig.~\ref{fig:intact-pipeline} shows the data‑flow.

\begin{figure}[H]
\centering
\resizebox{\linewidth}{!}{%
\begin{tikzpicture}[
  node distance=0.55cm and 0.65cm,
  box/.style={draw, rounded corners=2pt, align=center, font=\small, inner sep=4pt, minimum height=0.75cm},
  phase/.style={box, fill=myblue!18},
  data/.style={box, fill=mygray},
  arrow/.style={-{Stealth[length=2.2mm]}, thick, black!75}
]
\node[data] (live) {Live Sensing\\$\bar{V}_{\mathrm{o}},\,R_O,\,R_V$};
\node[data, right=of live] (predmod) {Travel‑Time Forecaster\\\texttt{LSTAN\_GERPE} + Driver Model};
\node[data, right=of predmod] (vpred) {Predicted Speed $\bar{V}_{\mathrm{f}}$};
\node[phase, below=0.9cm of live, xshift=0.8cm] (p1) {Phase 1\\Congestion Detection (TCD)};
\node[phase, right=of p1] (p2) {Phase 2\\Vehicle Selection (RVS)};
\node[phase, right=of p2] (p3) {Phase 3\\Alternative Route Allocation (ARA)};

\draw[arrow] (live) -- (predmod);
\draw[arrow] (predmod) -- (vpred);
\draw[arrow] (live.south) -- ++(0,-0.35) -| (p1.north);
\draw[arrow] (p1) -- (p2);
\draw[arrow] (p2) -- (p3);
\draw[arrow] (vpred.south) |- (p3.north);
\draw[arrow] (live.south) -- ++(0,-0.35) -| (p3.north);
\end{tikzpicture}%
}
\caption{Intact phase of the Selective-Rerouter: sensing, forecasting, and three‑phase selective control. TCD: traffic congestion detection, RVS: rerouted‑vehicle selection, and ARA: alternative‑route allocation.}
\label{fig:intact-pipeline}
\end{figure}
\FloatBarrier

\subsubsection{Phase 1: Traffic Congestion Detection (TCD)}
\label{pha:tcd}
Based on occupancy $R_O(s,\tau)$ and normalized speed $R_V(s,\tau)$ (Section~\ref{sec:system_model}), a segment $s$ is marked congested under hard thresholds $(\delta_O,\delta_V)=(0.55,0.45)$: $R_O(s,\tau) > \delta_O$ and $1-R_V(s,\tau) > \delta_V$. If hard condition fails, soft‑threshold rule triggers when $R_O(s,\tau) > \delta_O^{\mathrm{soft}}$, $1-R_V(s,\tau) > \delta_V^{\mathrm{soft}}$ and occupancy increases $R_O(s,\tau)>R_O(s,\tau-T)$, with $(\delta_O^{\mathrm{soft}},\delta_V^{\mathrm{soft}})=(0.35,0.35)$. Composite congestion severity score:
\begin{align}
R_S(s,\tau) &= R_O(s,\tau) \cdot \Psi_{ro} \nonumber \\
&+ \big(1-R_V(s,\tau)\big) \cdot \big(1-\Psi_{ro}\big).
\label{eq: R_S}
\end{align}
$\Psi_{ro}\in[0,1]$ weights occupancy contribution. Boolean predicate $isCongested(s,\tau)$ encodes hard‑or‑soft triggering. The set of congested segments:
\begin{equation}\label{eq: C_tau}
\mathcal{C}(\tau)=\bigl\{(s,R_S(s,\tau))\bigm| s\in\mathcal{S},\; isCongested(s,\tau)\bigr\}.
\end{equation}
Phase‑2 uses per‑segment severity $R_S(s,\tau)$ to configure upstream intervention depth.

\subsubsection{Phase 2: Rerouted‑Vehicle Selection (RVS)}
\label{subsec:rvs}
For each congested segment $s_c\in\mathcal{C}(\tau)$, upstream hop depth adapts to congestion severity and clips within bounds $[\theta_{ur}^{\min},\theta_{ur}^{\max}]$:
\begin{equation}\label{eq: theta_ur}
\theta_{ur}(s_c,\tau)
=\mathrm{clip}\Bigl(
\mathrm{round}\bigl(e^{a_{ur}\,R_S(s_c,\tau)+b_{ur}}\bigr),\,
[\theta_{ur}^{\min},\theta_{ur}^{\max}]
\Bigr).
\end{equation}
Larger $R_S(s_c,\tau)$ expands intervention scope. Standalone tests fix $\theta_{ur}=9$. Four‑corner evaluations apply severity‑dependent formula. Ablation Proposed-fixed~$\theta_{ur}$ uses constant value $9$.

Let $S_{ur}^{(i)}(s_c,\tau)$ denote $i$‑hop upstream segments of $s_c$. Total upstream topological window:
\begin{equation}\label{eq: SurT}
S_{ur}^{\mathrm{tot}}(s_c,\tau) = \bigcup_{i=1}^{\theta_{ur}(s_c,\tau)} S_{ur}^{(i)}(s_c,\tau).
\end{equation}
Base candidate pool collects vehicles within this window:
\begin{equation}\label{eq: S_rvT}
S_{rv}^{\mathrm{base}}(s_c,\tau)
=\bigl\{ v_r \,\big|\, s(v_r,\tau)\in S_{ur}^{\mathrm{tot}}(s_c,\tau) \bigr\}.
\end{equation}
We further add approaching vehicles outside this window whose remaining route will hit bottleneck $s_c$ within $\theta_{\mathrm{app}}$ hops:
\begin{equation}\label{eq: S_app}
\begin{aligned}
S_{\mathrm{app}}(s_c,\tau)
=\bigl\{&v \,\big|\,
s_c\in r(v,\tau),\;
1\le \mathrm{hops}(v,s_c)\le\theta_{\mathrm{app}},\\
&\quad s(v,\tau)\notin S_{ur}^{\mathrm{tot}}(s_c,\tau)
\bigr\}.
\end{aligned}
\end{equation}
Per‑bottleneck candidate set merges two groups:
\begin{equation}\label{eq: S_rv_final}
S_{rv}^{\mathrm{tot}}(s_c,\tau)=S_{rv}^{\mathrm{base}}(s_c,\tau)\cup S_{\mathrm{app}}(s_c,\tau).
\end{equation}
Duplicate vehicle entries are deduplicated, preserving minimal hop‑count metric for subsequent sorting. Finally merge results across all congested segments to obtain global candidate pool $S_{rv}^{\mathrm{tot}}(\tau)$.

\subsubsection{Phase 3: Alternative‑Route Allocation (ARA)}
\label{subsec:phase3_ara}
Candidates inside $S_{rv}^{\mathrm{tot}}(\tau)$ are sorted by ascending hop index and descending residual destination distance. For each vehicle, Yen’s $k$‑shortest‑path~\cite{10.2307/2629312} generates $N_{cr}$ alternative routes. Each candidate path is evaluated under multi‑objective normalized cost metrics: hybrid passing‑time $C_r^{p}$, path length $C^{d}$, route dissimilarity $C^{s}$, average segment occupancy $C^{o}$. Min‑max normalization is applied for every cost component:
\begin{equation}\label{eq: cost_norm}
\hat{C}^{(\zeta)}(r_k)
=\frac{C^{(\zeta)}(r_k)-\min_{r\in S_{cr}}C^{(\zeta)}(r)}
{\max_{r\in S_{cr}}C^{(\zeta)}(r)-\min_{r\in S_{cr}}C^{(\zeta)}(r)+\varepsilon_{\mathrm{den}}},
\end{equation}
$\varepsilon_{\mathrm{den}}>0$ prevents division‑by‑zero. Weighted composite cost selects output route:
\begin{equation}\label{eq: rhat}
\hat{r}(v_r,\tau)
=\arg\min_{r_k\in S_{cr}}
\sum_{\zeta\in\{p,d,s,o\}} w_{\zeta}\,\hat{C}^{(\zeta)}(r_k).
\end{equation}
Weights $w_{\zeta}$ are fixed hyper‑parameters (Table~\ref{tab:params}). For uncommitted civilians, the selected route replaces residual plan if it yields improvement. Special gateway‑sequence constraints are enforced for vehicles assigned to bundle carriage. Otherwise, the original route is retained. After route issuance, segment occupancy footprints are updated for subsequent decision steps.

Horizon‑dependent blending weight:
\begin{equation}\label{eq: hybrid_alpha}
\alpha(h)=\max\big(0,\alpha_0-h\cdot\alpha_{\mathrm{dec}}\big).
\end{equation}
Proposed-LIVE ablation sets $\alpha(h)\equiv 1$. Timing recursion computes arrival time and forecast horizon index $h_j$ for each path segment.

\subsubsection{Runtime Loop}
\label{subsec:main_operation}
At each decision boundary $\tau=nT$, Selective‑Rerouter computes congested‑segment set $\mathcal{C}(\tau)$. When non‑empty, Phase 2 vehicle‑selection and Phase3 route‑allocation execute following Fig.~\ref{fig:intact-pipeline}. Issued routes overwrite $r(v,\tau)$, altering vehicle membership for candidate pools in future timesteps. Cost-announcements are re-minimised and re-published every $T_{\mathrm{ref}}$ steps. A corner cache admits one only when it is strictly smaller, as required in Section~\ref{sec:boundary}. Upon receiving gateway-arrival events from \emph{corner} subregions, deferred civilian-carrier assignment executes under the deferred-binding constraint of Section~\ref{sec:boundary}.

\section{Performance Evaluation}
\label{sec:exp}
\setcounter{secnumdepth}{3}
\subsection{Simulation Setup and Evaluation Ground Rules}
\label{sec:setup}
\label{sec:paradigm}
The objective and the feasibility constraints are Section~\ref{sec:objective} and Section~\ref{sec:boundary}. This subsection fixes the protocol, the scores, and the admission rule used by every comparison below, and then the shared road-network setup those comparisons use. Section~\ref{sec:selch} is the proposed policy inside that solution space.

\subsubsection{Co-Simulation Evaluation Protocol}
\label{sec:evalprotocol}
This protocol scores any rerouting policy inside the solution space of Section~\ref{sec:boundary}. It fixes which module stays unchanged, which primitive a comparison may replace, and the order in which the intact side and a damaged corner meet. Fig.~\ref{fig:cosim-panels} in Section~\ref{sec:flow} is the data flow. The interface is Section~\ref{sec:interface}. The DTN-UAV Forwarder is Section~\ref{sec:dtn}. The \textsc{Reroute} primitive filled by the Selective-Rerouter is Algorithm~\ref{alg:reroute}.

Each global evaluation step follows a fixed execution order: the DTN-UAV Forwarder is updated first, followed by the candidate rerouting policy, and finally the thin-boundary interface synchronization. Only the \textsc{Reroute} primitive is replaced across candidate policies. \textsc{DTN-UAV-Step} and \textsc{Interface} stay fixed. The primitive \textsc{DTN-UAV-Step} is the fixed corner update in Section~\ref{sec:dtn}. It encodes the internal update logic of the DTN-UAV Forwarder. It consumes the cost-announcement of \eqref{eq:quote} and does not recompute it. \textsc{Interface} publishes the beacon of Section~\ref{sec:interface} and applies \eqref{eq:cfree} after an arrival $(g,u)$. A candidate that cannot emit \eqref{eq:cbeacon}, or that names a civilian before \eqref{eq:carrival}, is not a feasible policy under this protocol.

\begin{algorithm}[!t]
\LinesNumbered
\KwIn{fixed primitives \textsc{DTN-UAV-Step} and \textsc{Interface}, candidate primitive \textsc{Reroute}}
\KwOut{receipts on the named oscillators}
\While{a civilian remains in the intact area, or an inbound bundle is still waiting}{
  \textsc{DTN-UAV-Step}()\;
  \textsc{Reroute}($\mathcal{V}$)\;
  \textsc{Interface}()\;
}
\caption{Global evaluation main loop for our co-simulation protocol. Only the \textsc{Reroute} primitive is substituted for candidate rerouting policies.}
\label{alg:platform}
\end{algorithm}

\subsubsection{Performance Metrics}
\label{sec:metrics}
Five end-to-end scores are defined below. $D_{\mathrm{gw}}$, $\bar{\Delta}_{\mathrm{gw}}$, and $A_{\mathrm{veh}}$ inspect the handover and are not substitutes for them. Corner-local times are not among the five. Arrival at a gateway is not a receipt: the copy must arrive at the named oscillator $o^{\star}_m$.

$D_{\mathrm{osc}}$ counts a bundle once, when its copy arrives at destination oscillator $o^{\star}_m$. The last hop is the destination gateway when \eqref{eq:named} holds, and otherwise that corner's UAV. Arrival at the gateway is not a receipt. $D_{\mathrm{osc}}/M$ is the share of the $M$ objects born. $T_m$ is the time from birth to receipt, over received objects only. $\bar{T}_{\mathrm{obj}}=T_{\Sigma}/D_{\mathrm{osc}}$ when $D_{\mathrm{osc}}>0$, and $T_{\mathrm{obj}}^{\max}$ is the longest receipt. $\bar{t}_{\mathrm{civ}}$ is the mean duration of completed civilian trips.

$D_{\mathrm{gw}}$ counts source-gateway handovers before $T_{\mathrm{TTL}}$. Here $\mathrm{gw}$ denotes a gateway.
\begin{equation}
D_{\mathrm{gw}}=\sum_{m\in\mathcal{M}} Y_m ,
\label{eq:Dgw}
\end{equation}
where $Y_m=1$ when that handover occurs. $\Delta_m$ is the published extra time of the receiving gateway at the drop, the cost-announcement recorded then, and is not the driver scale $\kappa(v_r,s)$. $\bar{\Delta}_{\mathrm{gw}}$ is the mean of those cost-announcements. A drop with no finite cost-announcement is reported separately and is not entered as zero. $A_{\mathrm{veh}}$ counts both-gateway routes written after $(g,u)$. Section~\ref{sec:sub1val} compares forwarders inside one corner and is not a substitute for $D_{\mathrm{osc}}$, $\bar{T}_{\mathrm{obj}}$, or $\bar{t}_{\mathrm{civ}}$.

\begin{table}[H]
\caption{Scores of a run. The first block is the end-to-end comparison \eqref{eq:objmin}--\eqref{eq:objmax}. The second block inspects the handover and is not a substitute for the first.}
\label{tab:scores}
\centering
\footnotesize
\renewcommand{\arraystretch}{1.08}
\begin{tabular}{@{}llp{0.42\columnwidth}@{}}
\toprule
Score & Better & Reading \\
\midrule
$D_{\mathrm{osc}}$ & larger & Receipts at $o^{\star}_m$ \\
$D_{\mathrm{osc}}/M$ & larger & Share of the demand \\
$\bar{T}_{\mathrm{obj}}$ & smaller & Mean birth-to-receipt time \\
$T_{\mathrm{obj}}^{\max}$ & smaller & Longest receipt \\
$\bar{t}_{\mathrm{civ}}$ & smaller & Mean completed civilian trip \\
\midrule
$D_{\mathrm{gw}}$ & --- & Source-gateway handovers before $T_{\mathrm{TTL}}$ \\
$\bar{\Delta}_{\mathrm{gw}}$ & --- & Mean cost-announcement at those handovers \\
$A_{\mathrm{veh}}$ & --- & Both-gateway routes written \\
\bottomrule
\end{tabular}
\end{table}
\FloatBarrier

\subsubsection{Baseline Eligibility}
\label{sec:eligibility}
\label{sec:related-dtn}
\label{sec:families}
In every end-to-end run the following stay fixed: one copy per bundle, the UAV geofence, the tour stop and the in-radio shift, the radio range, the oscillator patrols, the bundle births, the lifetime $T_{\mathrm{TTL}}$, and the two messages of Section~\ref{sec:interface}. The cost-announcement follows \eqref{eq:quote} in Section~\ref{sec:interface}. Section~\ref{sec:publishers} records the generation rule used in the reported runs. The only replacement is the rerouting policy on the intact side. The Selective-Rerouter is the proposed policy, and a replacement must still emit \eqref{eq:cbeacon}, must still wait for \eqref{eq:carrival} before naming a civilian, and must still apply \eqref{eq:cfree}. A gap among such runs is a gap among those policies. The DTN-UAV Forwarder is the same on every run.

Classical delay-tolerant routing protocols do not meet the deferred-binding and cost-announcement constraints of Section~\ref{sec:boundary}. A design that names a node or a vehicle before the bundle is in a gateway buffer, or that cannot consume \eqref{eq:quote}, is not an end-to-end peer. Epidemic, PRoPHET, and MaxProp rank replicas from contact history \cite{fall2003dtn,vahdat2000epidemic,lindgren2003prophet,burgess2006maxprop}. Ferries and UAV relays steer a dedicated node to create contacts \cite{zhao2004ferry}. First-gateway handover is an internal reference: it drops at the first gateway in radio and ignores \eqref{eq:quote}. Table~\ref{tab:position} records whether each design satisfies those feasibility constraints. The classical rows miss deferred binding and the cost-announcement, so they stay outside the end-to-end comparison. Epidemic, PRoPHET, and first-gateway handover remain corner-level references. The selective pipeline is not a row of that table. MaxProp is outside the corner check. A forwarder that obeys \eqref{eq:cbeacon}--\eqref{eq:cfree} and differs from Section~\ref{sec:dtn} is admissible as a later row.

\begin{table}[H]
\caption{The DTN-UAV Forwarder against delay-tolerant designs. Each row records whether that design satisfies the feasibility constraints of Section~\ref{sec:boundary}.}
\label{tab:position}
\centering
\footnotesize
\setlength{\tabcolsep}{4pt}
\begin{tabular}{@{}lcccc@{}}
\toprule
 & DTN & Ferry & UAV & Sub-mod. \\
\midrule
Single-copy store-carry-forward & yes & yes & yes & one copy \\
Layer of the new rule & route & route & motion & application \\
Civilian trip is the carrier & no & no & no & after handover \\
Dedicated extra fleet & no & yes & aerial & no \\
Intact-area cost-announcement & no & no & no & yes \\
Follows a lower cost-announcement & no & no & no & yes \\
Vehicle named before handover & no & yes & no & no \\
Both gateways, one written route & no & no & no & after handover \\
\bottomrule
\end{tabular}
\end{table}
\FloatBarrier

Section~\ref{sec:performance_evaluation} isolates the Selective-Rerouter on undamaged districts. Section~\ref{sec:sub1val} then checks the DTN-UAV Forwarder on one corner, with a simulated cost-announcement and no intact fleet. Four-corner runs after that check replace only the rerouting policy on the intact side.

The end-to-end family keeps both components. The procedure is the Selective-Rerouter or one of NRR, ReFOCUS+, FCR, CCRGS, Du-GAQ, and CAIE-TT-Scoped. Each replacement still publishes a cost-announcement and waits for $(g,u)$ before naming a civilian. Scores are the first block of Table~\ref{tab:scores}. Ablations that vary one term of the same policy are not a second family. The standalone family loads no damaged corner, so \eqref{eq:objmax} is undefined. Its scores are mean civilian duration and mean executed route changes. Corner-check counts do not enter the end-to-end scores.

\subsubsection{Network and Shared Parameters}
\begin{figure}[!t]
\centering
\includegraphics[width=\columnwidth]{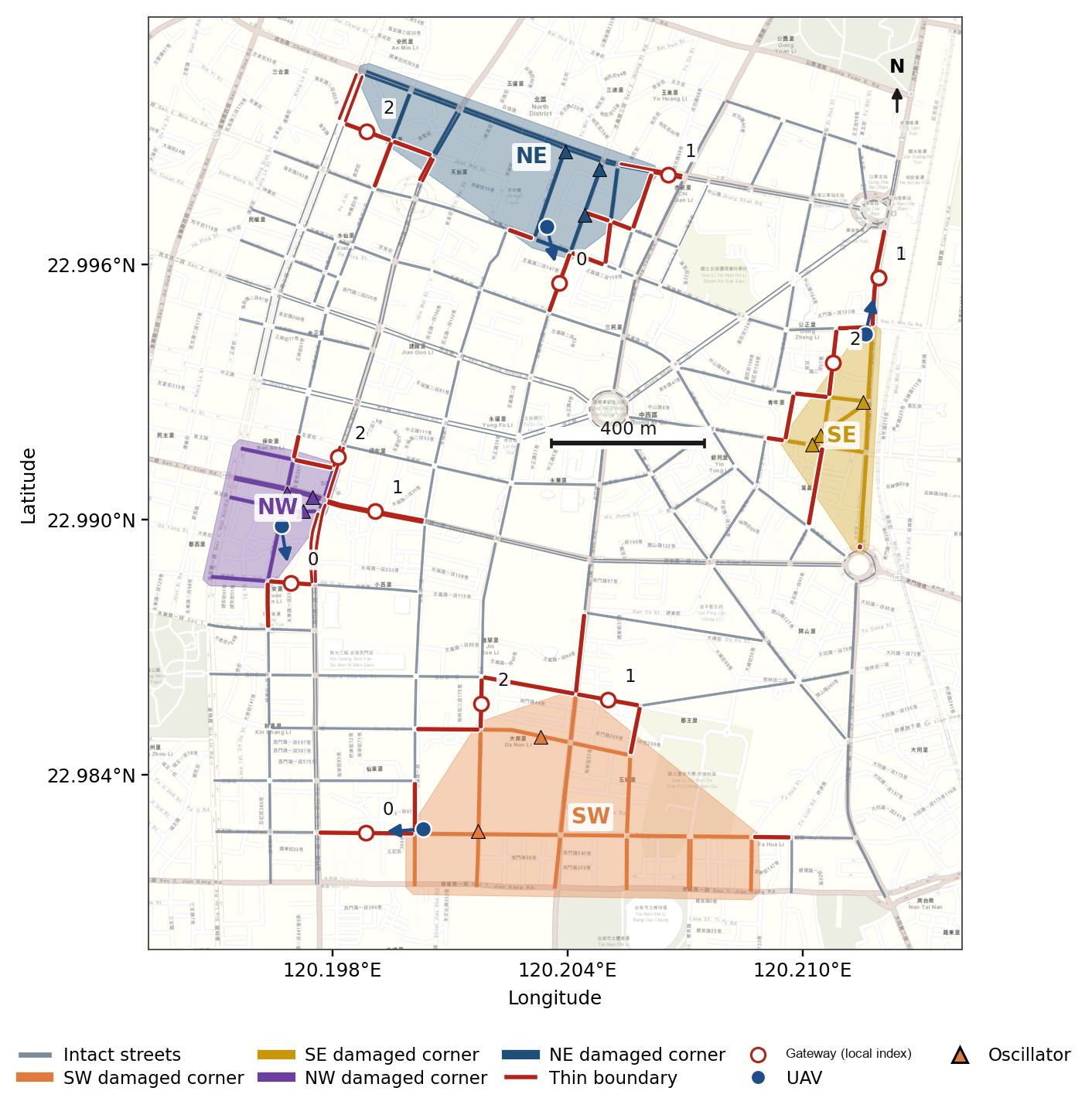}
\caption{Road-network topology of the West Central District of Tainan, used to instantiate the hypothetical post-disaster scenario. Colored regions are artificially injected corners, pale lines are the intact SUMO streets, and red edges mark the thin boundary. The partition is artificial and does not represent a real-world disaster at this location. Basemap: Esri, HERE, Garmin, \copyright\ OpenStreetMap contributors.}
\label{fig:tainan-map}
\end{figure}

In our simulation evaluation, we adopt $|\mathcal{Q}|=4$ corner subregions, $n_g=3$ gateways per corner, and bundle expiry $T_{\mathrm{TTL}}=10000\,\mathrm{s}$.
We adopt the road-network topology of the West Central District of Tainan from OpenStreetMap to instantiate such hypothetical post-disaster scenarios. Network partitions and road impairments are artificially injected for simulation purposes and do not represent real-world disasters that occurred in this location. Every run in this section uses that topology, drawn in Fig.~\ref{fig:tainan-map}. Two SUMO setups score the Selective-Rerouter on that network~\cite{lopez2018sumo}. The standalone setup uses the pale streets of this map, with the four corners left intact and with no cross-corner bundles. The four-corner setup artificially injects the partition on the same map: the colored corners, the red thin boundary, and the gateways, UAVs, and oscillators drawn there, with the DTN-UAV Forwarder held fixed. The dual graph of these streets is divided into $N_z=14$ geographic zones. Four-corner civilian trips are released over $10$ demand regions on the intact network, and each end-to-end table names the civilian and object counts of that run. Each corner has three gateways, one UAV, and three oscillators. Each bundle from oscillator $i$ in corner $q$ is addressed to the gateway and the ground-patrol oscillator sharing index $i$ in the remote corner. Gateways broadcast a scalar cost and do not disclose a civilian-vehicle identifier. The corner-level check uses the southwest corner of this map. Congestion detection uses hard thresholds $(\delta_O,\delta_V)=(0.55,0.45)$ and soft thresholds $(\delta_O^{\mathrm{soft}},\delta_V^{\mathrm{soft}})=(0.35,0.35)$. Severity weight $\Psi_{ro}=0.5$ combines occupancy-speed metrics for congested-segment ranking. The base rerouting decision period is $T=300\,\mathrm{s}$. The default forecast step is $T_h=300\,\mathrm{s}$, for the standalone runs and for the four-corner mainline. The $450/720$ and $900/360$ loads use a forecast trained and evaluated at $T_h=60\,\mathrm{s}$. The $1800/720$ load evaluates the $300\,\mathrm{s}$ mainline weights at $T_h=60\,\mathrm{s}$. Cost-announcement beacons in the end-to-end runs are re-minimised and re-published every $T_{\mathrm{ref}}=10$ simulation steps. A corner cache writes a re-published value only when it is strictly smaller. Speed floor $\varepsilon_{\mathrm{spd}}=10^{-4}\,\mathrm{m/s}$ is used for numerical stability to avoid division-by-zero. Table~\ref{tab:params} lists the hyperparameters. The corner rules are Section~\ref{sec:corner}. Runs that scored cost-announcements from the current mean speed alone use a different rerouting policy and are absent from the tables below.

\setcounter{secnumdepth}{2}
\subsection{Standalone Evaluation on Undamaged Urban Districts}
\label{sec:performance_evaluation}
This block excludes damaged corners and cross-corner bundles, so the score is civilian travel time (Section~\ref{sec:metrics}). Reroute count RR is the mean number of executed route changes. It is an auxiliary metric and should not be minimized on its own.

\subsubsection{Simulation Scenarios and Parameter Sweep}
\label{sec:sim_settings}
\label{subsec:sim_params}
\label{subsec:theta_ur_param}
\begin{figure}[!t]
\centering
\includegraphics[width=0.92\columnwidth]{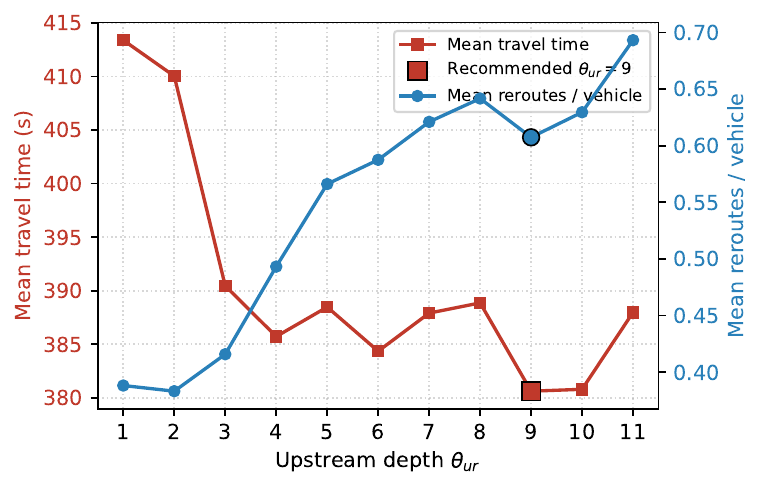}
\caption{Upstream-depth sweep on the recommended stack ($8000$ vehicles). The highlighted marker denotes $\theta_{ur}{=}9$.}
\label{fig:theta_ur_sweep}
\end{figure}

The intact network is the pale street grid of Fig.~\ref{fig:tainan-map}, taken from the West Central District of Tainan road-network topology: $204$ intersections, $561$ segments, and $77.3$\,km, divided into $14$ traffic zones and $20$ origin--destination pairs. Demand is released at the start. The representative load is $8000$ vehicles. The same Selective-Rerouter is also run at $16000$ and $20000$. The recommended stack freezes the LSTAN-GERPE~\cite{WangYang2025LSTAN} encoder and fine-tunes only the ranking head (Table~\ref{tab:params}). Fig.~\ref{fig:theta_ur_sweep} sweeps the upstream window at $8000$ vehicles. A window that is too small under-covers the bottleneck, and excessive depth pulls in distant vehicles and creates secondary congestion. The operating point is $\theta_{ur}{=}9$.

\subsubsection{Ablation and Baseline Comparisons}
\label{subsec:ablation_study}
\label{subsec:baseline_comparison}
\label{subsec:mechanism_analysis}
Leave-one-out ablation on the matched $8000$-vehicle set (Table~\ref{tab:ablation_standalone}) shows that hybrid live--forecast costing supplies the dominant travel-time gain. The other Selective-Rerouter terms, and the choice of forecast head, produce smaller differences. Against the literature rerouters, Selective-Rerouter keeps the lowest mean travel time at $8000$, $16000$, and $20000$ vehicles, and the gap to same-scope live Dijkstra widens from $11.0\,\mathrm{s}$ to $115.9\,\mathrm{s}$ and $148.6\,\mathrm{s}$. Relative to the no-reroute floor of $1387.1\,\mathrm{s}$ at $8000$ vehicles, Selective-Rerouter cuts mean travel time by $72.6\%$.

\begin{table}[!t]
\centering
\fontsize{6.5}{7.6}\selectfont
\renewcommand{\arraystretch}{0.92}
\setlength{\tabcolsep}{1.5pt}
\caption{Simulation parameters. ``Four-corner'': artificially partitioned topology with the DTN-UAV Forwarder. ``Standalone'': intact-only topology. An em dash (---) marks parameters that do not apply.}
\label{tab:params}
\begin{tabular*}{\columnwidth}{@{\extracolsep{\fill}}ll@{}}
\begin{tabular}[t]{@{}>{\raggedright\arraybackslash}p{0.25\columnwidth}>{\raggedright\arraybackslash}p{0.21\columnwidth}@{}}
\multicolumn{2}{@{}l}{\textit{Shared parameters (A)}} \\
Rerouting period $T$ & $300\,\mathrm{s}$ \\
Candidate budget $N_{cr}$ & $7$ \\
Vehicle length $\bar{L}_v$, spacing $L_{sd}$ & $5\,\mathrm{m}$, $2.5\,\mathrm{m}$ \\
Thresholds $(\delta_O,\delta_V)$ & $(0.55,0.45)$ \\
Soft thresholds & $(0.35,0.35)$ \\
\multicolumn{2}{@{}l}{$(\delta_O^{\mathrm{soft}},\delta_V^{\mathrm{soft}})$} \\
Severity weight $\Psi_{ro}$ & $0.5$ \\
Depth-map $(a_{ur},b_{ur})$ & $(5.37,-3.07)$ \\
Clip bounds & $[2,10]$ \\
\multicolumn{2}{@{}l}{$[\theta_{ur}^{\min},\theta_{ur}^{\max}]$} \\
Approaching hops $\theta_{\mathrm{app}}$ & $2$ \\
Blend $(\alpha_0,\alpha_{\mathrm{dec}})$ & $(0.75,0.12)$ \\
Feature dimension $F$ & $2$ (speed, occupancy) \\
Rank loss $\lambda_{\mathrm{rank}}$ & $0.015$ \\
Min path gap $t_{\min}$ & $10\,\mathrm{s}$ \\
\midrule
\multicolumn{2}{@{}l}{\textit{Setup-dependent}} \\
Four-corner $(w_p,w_d,w_s,w_o)$ & $(0.35,0.15,$\newline $0.15,0.35)$ \\
Standalone $(w_p,w_d,w_s,w_o)$ & $(0.45,0.20,$\newline $0.15,0.20)$ \\
$\theta_{ur}$, four-corner & \eqref{eq: theta_ur} \\
$\theta_{ur}$, standalone & $9$ \\
$T_h$, default & $300\,\mathrm{s}$ \\
$T_h$, $450/720$, $900/360$ & $60\,\mathrm{s}$ \\
$T_h$, $1800/720$ & $60\,\mathrm{s}$ step, $300\,\mathrm{s}$ weights \\
\midrule
\multicolumn{2}{@{}l}{\textit{Four-corner only}} \\
Damaged fraction per island & $0.30$ \\
Corners $|\mathcal{Q}|$ & $4$, $\{0,1,2,3\}$ \\
Gateways per corner $n_g$ & $3$ \\
Oscillators per corner $|\mathcal{O}_q|$ & $3$ \\
Active UAV count & $|\mathcal{Q}|$ \\
\end{tabular}
&
\begin{tabular}[t]{@{}>{\raggedright\arraybackslash}p{0.25\columnwidth}>{\raggedright\arraybackslash}p{0.21\columnwidth}@{}}
\multicolumn{2}{@{}l}{\textit{Shared parameters (B)}} \\
Fine-tune batch / OD paths & $8$ OD / $32$ \\
Fine-tune learning rate & $5{\times}10^{-5}$ \\
$Bd$ sample count $n_s$ & $50$ \\
Congestion selector & on, selective \\
Speed floor $\varepsilon_{\mathrm{spd}}$ & $10^{-4}\,\mathrm{m/s}$ \\
Windows $(T_{\mathrm{in}},T_{\mathrm{out}})$ & $(12,6)$ \\
Geographic zones $N_z$ & $14$ \\
\midrule
\multicolumn{2}{@{}l}{\textit{Four-corner only, continued}} \\
Demand regions & $10$ \\
Radio range $R$ & $400\,\mathrm{m}$ \\
World-coordinate scale & $10000$ \\
Bundle lifetime $T_{\mathrm{TTL}}$ & $10000\,\mathrm{s}$ \\
Gateway-queue timeout & $10000\,\mathrm{s}$ \\
Civilian trip counts & $400$, $450$, $900$, $1800$, $2700$ \\
Bundle emission instants & $1000$--$8600\,\mathrm{s}$, step $400$ \\
Bundle count per trial & $350$, $360$, $720$, $1080$, $1440$, $2160$ \\
UAV speed & $100$ world units/step \\
Oscillator laps / depart. & $70$ / $0,1,2\,\mathrm{s}$ \\
Civilian depart. (mode~1) & $1$--$9800\,\mathrm{s}$ \\
Corridor / slack $(w_{\mathrm{cor}},\sigma)$ & $80\,\mathrm{m}$ / $40\,\mathrm{m}$ \\
Shift / margins $(s_{\mathrm{shift}},\eta_{\mathrm{mrg}},\varepsilon_{\mathrm{mrg}})$ & $160\,\mathrm{m}$ / $80\,\mathrm{m}$, $40\,\mathrm{m}$ \\
Tie margin $\delta_{\mathrm{tie}}$ & $30\,\mathrm{m}$ \\
Dwell / stall $(T_{\mathrm{dwell}},T_{\mathrm{stall}})$ & $60$ / $30$ steps \\
Preemption $T_{\mathrm{pre}}$ & $800\,\mathrm{s}$ (on oscillator) \\
Re-publish $T_{\mathrm{ref}}$ & every $10$ steps \\
Alert offset $T_{\mathrm{alert}}$ & $150\,\mathrm{s}$ \\
Test pool $N_{\mathrm{try}}$ & $32$ \\
Cost-announcement rule & Sec.~\ref{sec:publishers} \\
\end{tabular}
\end{tabular*}
\end{table}

Note that the ablation in Section~\ref{sec:performance_evaluation} is conducted in the standalone intact-district setting without bundles, to isolate the intrinsic traffic-level contribution of each Selective-Rerouter component. The leave-one-out ablation in Table~\ref{tab:ablation400} repeats similar component perturbations under full four-corner co-simulation, to quantify their joint impact on both civilian travel cost and cross-corner bundle-delivery performance. Results from the two settings are not interchangeable.

Table~\ref{tab:ablation_standalone} is that intact-district ablation. There is no damaged corner and no cross-corner bundles, so the only score is mean civilian travel time. The increase column is the change in mean travel time relative to the recommended Selective-Rerouter with $\theta_{ur}{=}9$ and ranking-head fine-tuning. Block~I changes one Selective-Rerouter term. Block~II keeps that Selective-Rerouter and replaces only the travel-time forecaster. Removing hybrid live--forecast costing (Proposed-LIVE) is the largest Block~I penalty. Both forecast-head swaps remain smaller than that penalty.

\begin{table}[H]
\centering
\caption{Standalone ablation on the intact Tainan road-network topology ($8000$ vehicles, no bundles and no injected partition). Mean travel time and the increase versus the recommended Selective-Rerouter. Bold is the lowest mean. Underline is the second-lowest mean and the increase on that row.}
\label{tab:ablation_standalone}
\footnotesize
\setlength{\tabcolsep}{4pt}
\renewcommand{\arraystretch}{0.86}
\begin{tabular}{@{}lrr@{}}
\toprule
Configuration & Mean (s) & Increase (s) \\
\midrule
\multicolumn{3}{@{}l@{}}{\textit{Block I: Selective-Rerouter terms}} \\
\textbf{Proposed} & \textbf{380.6} & \textbf{---} \\
Proposed w/o $R_V$ & $400.3$ & $+19.7$ \\
Proposed-fixed~$\theta_{ur}$ & \rkii{385.7} & \rkii{+5.1} \\
Proposed-LIVE & $438.1$ & $+57.5$ \\
Proposed-time & $405.6$ & $+25.0$ \\
Proposed-noDB & $393.0$ & $+12.4$ \\
\midrule
\multicolumn{3}{@{}l@{}}{\textit{Block II: forecaster on the recommended stack}} \\
Proposed-LSTM (per-segment) & $388.2$ & $+7.6$ \\
Proposed-Huber & $400.8$ & $+20.2$ \\
\bottomrule
\end{tabular}
\end{table}
\FloatBarrier

\subsubsection{Computational Overhead}
\label{subsec:complexity}
On this graph ($N{=}561$, $N_{cr}{=}7$), Phase-3 computation takes $1.45\,\mathrm{s}$ at the representative selective set $|S_{rv}^{\mathrm{tot}}|{\approx}102$, $2.44\,\mathrm{s}$ at the observed peak $|S_{rv}^{\mathrm{tot}}|{=}198$, and $3.89\,\mathrm{s}$ at a stress load $|S_{rv}^{\mathrm{tot}}|{=}300$. Each of these fits inside the decision interval $T{=}300\,\mathrm{s}$. Path enumeration is the bottleneck. The batched forecast takes a fixed $0.31\,\mathrm{s}$ and does not grow with $|S_{rv}^{\mathrm{tot}}|$. Practical mitigations are caching of $k$-shortest paths, a smaller $N_{cr}$ under mild congestion, and bounded search. The $8000$-vehicle co-simulation, including SUMO, finishes on the order of two minutes.

\setlength{\intextsep}{4pt plus 1pt minus 1pt}
\setlength{\textfloatsep}{6pt plus 2pt minus 2pt}
\setlength{\floatsep}{4pt plus 1pt minus 1pt}

\subsection{Corner-Level Experiments}
\label{sec:sub1val}
The DTN-UAV Forwarder is checked on the southwest corner of Fig.~\ref{fig:tainan-map} first. That corner holds three oscillators, one UAV, and $600$ confidential objects, with no civilian trips. Tables~\ref{tab:zoneexit} and~\ref{tab:zoneproto} check it against the thin-boundary contract (Section~\ref{sec:eligibility}). Epidemic and PRoPHET are corner-level references only and are ineligible for the end-to-end comparison. Delivery is sensitive to radio range. Under this static simulated cost-announcement, changing the re-publication interval does not rename the gateway, because a repeated announcement is not strictly smaller and \eqref{eq:cacheupd} leaves the cached exit in place.

\begin{table}[H]
\centering
\caption{Corner-only checks of the fixed forwarder. $R$ is in meters and times are in seconds. $D_{\mathrm{gw}}$ counts arrivals out of $600$. The cost-announcement is straight-line time at $8\,\mathrm{m/s}$, and the horizon is $1800\,\mathrm{s}$. Means use objects that reach a gateway. The expiry rows discard a stored cost-announcement after $T_{\mathrm{ref}}$ steps.}
\label{tab:zoneexit}
\footnotesize
\setlength{\tabcolsep}{3pt}
\renewcommand{\arraystretch}{0.96}
\begin{tabular*}{\columnwidth}{@{\extracolsep{\fill}}lrrrrrrr@{}}
\toprule
Policy & $R$ & $T_{\mathrm{ref}}$ & $D_{\mathrm{gw}}$ & Quoted & Wrong & $\bar{T}_{\mathrm{gw}}$ & $\bar{\Delta}_{\mathrm{gw}}$ \\
\midrule
Quoted gateway & $400$ & $10$ & $600$ & $600$ & $0$ & $386.9$ & $125.4$ \\
First gateway & $400$ & $10$ & $600$ & $303$ & $297$ & $300.2$ & $137.1$ \\
No UAV & $400$ & $10$ & $600$ & $600$ & $0$ & $0.7$ & $125.4$ \\
Quoted gateway & $800$ & $10$ & $600$ & $600$ & $0$ & $299.2$ & $125.4$ \\
Quoted gateway & $200$ & $10$ & $10$ & $10$ & $0$ & $57.2$ & $115.9$ \\
Quoted gateway & $80$ & $10$ & $5$ & $5$ & $0$ & $91.4$ & $109.2$ \\
Quoted, expires & $400$ & $1$ & $600$ & $600$ & $0$ & $386.9$ & $125.4$ \\
Quoted, expires & $400$ & $10$ & $600$ & $600$ & $0$ & $386.9$ & $125.4$ \\
Quoted, expires & $400$ & $60$ & $600$ & $600$ & $0$ & $386.9$ & $125.4$ \\
\bottomrule
\end{tabular*}
\end{table}

\begin{table}[H]
\centering
\caption{Corner-only forwarding on the same $600$ objects. The Quoted-gateway row is the DTN-UAV Forwarder from Table~\ref{tab:zoneexit}. Quoted and Wrong count handovers at the minimum cost-announcement gateway and at a different gateway.}
\label{tab:zoneproto}
\footnotesize
\setlength{\tabcolsep}{3.5pt}
\renewcommand{\arraystretch}{0.96}
\begin{tabular*}{\columnwidth}{@{\extracolsep{\fill}}lrrrrr@{}}
\toprule
Forwarder & $R$ (m) & $D_{\mathrm{gw}}/600$ & Quoted & Wrong & $\bar{T}_{\mathrm{gw}}$ (s) \\
\midrule
Quoted gateway & $400$ & $600$ & $600$ & $0$ & $386.9$ \\
Epidemic & $400$ & $600$ & $267$ & $333$ & $0.7$ \\
PRoPHET & $400$ & $600$ & $600$ & $0$ & $0.7$ \\
Quoted gateway & $200$ & $10$ & $10$ & $0$ & $57.2$ \\
Epidemic & $200$ & $600$ & $267$ & $333$ & $0.7$ \\
PRoPHET & $200$ & $600$ & $600$ & $0$ & $38.4$ \\
\bottomrule
\end{tabular*}
\end{table}

\subsection{End-to-End Co-Simulation Experiments}
\label{sec:e2e}
\label{sec:hybrid400}
The four-corner runs follow the artificially injected partition in Fig.~\ref{fig:tainan-map} and the corner check. Every end-to-end trial follows the evaluation ground rules in Section~\ref{sec:paradigm}. The DTN-UAV Forwarder stays fixed, and only the rerouting policy on the intact side changes.

\subsubsection{Representative Full End-to-End Hybrid-Cost Trial}
Table~\ref{tab:hybrid400} summarizes aggregate outcomes for one complete hybrid-cost run with $400$ civilians and $350$ confidential objects under disabled teleport. In this trial, all civilian trips finish and every cross-corner object achieves full end-to-end delivery to its target oscillator. Driven by cost-announcement-aware gateway selection that prefers lower-cost gateways, bundle hand-off load is unevenly distributed, and some gateways produce zero outgoing bundle traffic. This imbalanced gateway utilization is an expected behavior of the cost-driven selection mechanism and does not degrade overall delivery performance.

\begin{table}[H]
\caption{Hybrid-cost run on the Tainan road-network topology with an artificially injected four-corner partition: $400$ civilians, $350$ confidential objects, teleport disabled.}
\label{tab:hybrid400}
\centering
\footnotesize
\setlength{\tabcolsep}{3pt}
\renewcommand{\arraystretch}{0.92}
\begin{tabular*}{\columnwidth}{@{\extracolsep{\fill}}%
  >{\raggedright\arraybackslash}p{0.26\columnwidth}r%
  >{\raggedright\arraybackslash}p{0.26\columnwidth}r@{}}
\toprule
Quantity & Value & Quantity & Value \\
\midrule
$D_{\mathrm{gw}}$ & $350/350$ & $D_{\mathrm{osc}}$ & $350$ \\
Drops with no finite cost-announcement & $0$ & $\bar{t}_{\mathrm{civ}}$ & $848.4\,\mathrm{s}$ \\
Expired & $0$ & Mean civilian time loss & $489.7\,\mathrm{s}$ \\
$A_{\mathrm{veh}}$ accepted / rejected & $350/0$ & $T_{\Sigma}$ & $236830\,\mathrm{s}$ \\
Mean scored extra time & $345.4\,\mathrm{s}$ & $\bar{T}_{\mathrm{obj}}$ & $676.7\,\mathrm{s}$ \\
Mean accepted extra time & $443.6\,\mathrm{s}$ & $T_{\mathrm{obj}}^{\max}$ & $1128\,\mathrm{s}$ \\
Civilians that carried an object & $350$ & Civilians arrived & $400/400$ \\
Destination-gateway arrivals & $350$ & Stop time & $1512\,\mathrm{s}$ \\
\bottomrule
\end{tabular*}
\end{table}

\subsubsection{Ablation Study under Multi-Civilian-Fleet Loads}
\label{sec:ablation400}
\label{sec:ablation900}
Table~\ref{tab:ablation400} removes one Selective-Rerouter term at two demands. The $400/350$ block uses the default $T_h=300\,\mathrm{s}$ stack and is the hybrid-cost run in Table~\ref{tab:hybrid400}. The $1800/720$ block evaluates the same $300\,\mathrm{s}$ weights at $T_h=60\,\mathrm{s}$, matching that row of Table~\ref{tab:baseline900}. Removing hybrid live--forecast costing raises both means at both demands. On both demands, Selective-Rerouter keeps the lowest civilian mean and the lowest object mean. At $1800/720$, driver scaling is the only row that misses receipts, $675$ of $720$, and time-only scoring has a shorter longest receipt. The standalone ablation, with no bundles, is Table~\ref{tab:ablation_standalone} in Section~\ref{sec:performance_evaluation}.

\begin{table}[H]
\scriptsize
\setlength{\tabcolsep}{1.5pt}
\renewcommand{\arraystretch}{0.85}
\caption{Leave-one-out ablation at two demands. Times are in seconds. $400/350$ is $400$ civilians and $350$ objects. $1800/720$ is $1800$ civilians and $720$ objects. Each row is one configuration. The two column groups are the two demands. Bold is the best value in that block: the lowest time, or the highest $D_{\mathrm{osc}}$. Underline is the second-best distinct value. A tie is marked on every such row.}
\label{tab:ablation400}
\begin{tabular*}{\columnwidth}{@{\extracolsep{\fill}}lrrrr@{\hspace{3pt}}rrrr@{}}
\toprule
& \multicolumn{4}{c}{$400/350$} & \multicolumn{4}{c}{$1800/720$} \\
\cmidrule(lr){2-5}\cmidrule(lr){6-9}
Configuration & \makecell[r]{$\bar{t}_{\mathrm{civ}}$\\(s)} & \makecell[r]{$\bar{T}_{\mathrm{obj}}$\\(s)} & \makecell[r]{$T_{\mathrm{obj}}^{\max}$\\(s)} & $D_{\mathrm{osc}}$ & \makecell[r]{$\bar{t}_{\mathrm{civ}}$\\(s)} & \makecell[r]{$\bar{T}_{\mathrm{obj}}$\\(s)} & \makecell[r]{$T_{\mathrm{obj}}^{\max}$\\(s)} & $D_{\mathrm{osc}}$ \\
\midrule
Proposed & \textbf{848.4} & \textbf{676.7} & \textbf{1128} & \textbf{350} & \textbf{1014.2} & \textbf{1098.8} & $1982$ & \textbf{720} \\
Proposed-LIVE & $1163.3$ & $770.0$ & $1532$ & \textbf{350} & $1439.1$ & $1539.0$ & $2657$ & \textbf{720} \\
Proposed w/o $R_V$ & $1068.0$ & \rkii{694.2} & $1461$ & \textbf{350} & $1157.3$ & $1253.9$ & $2030$ & \textbf{720} \\
Proposed-fixed~$\theta_{ur}$ & \rkii{1045.7} & $700.5$ & $1455$ & \textbf{350} & \rkii{1108.2} & \rkii{1176.2} & \rkii{1956} & \textbf{720} \\
Proposed-time & $1071.2$ & $697.9$ & \rkii{1407} & \textbf{350} & $1109.3$ & $1212.3$ & \textbf{1940} & \textbf{720} \\
Proposed-noDB & $1112.4$ & $744.2$ & $1431$ & \textbf{350} & $2135.3$ & $3117.8$ & $9512$ & \rkii{675} \\
\bottomrule
\end{tabular*}
\end{table}

\subsubsection{Baseline Comparison across Multiple Demand Configurations}
\label{sec:baseline400}
\label{sec:baseline900}
Table~\ref{tab:baseline900} compares Selective-Rerouter with NRR~\cite{7433412}, ReFOCUS+~\cite{8901448}, FCR~\cite{MatsuiYoshihiro2024FCR}, CCRGS~\cite{Dong2026CCRGS}, Du-GAQ~\cite{Du2024CACAIE}, and CAIE-TT-Scoped~\cite{WlodarczykSaber2026CAIE} on eight mode-2 demands. The literature rows keep their published configurations and were not retrained. Selective-Rerouter uses the hard gates $(\delta_O,\delta_V)=(0.55,0.45)$ and the soft gates $(\delta_O^{\mathrm{soft}},\delta_V^{\mathrm{soft}})=(0.35,0.35)$. The default forecast step is $T_h=300\,\mathrm{s}$. The $900/720$, $900/1080$, $1800/1440$, $2700/1440$, and $2700/2160$ rows use that step. $900/1080$ continues from the mainline weights, $1800/1440$ is the first scene-fit test, $2700/1440$ is the scale run on the mainline weights, and $2700/2160$ is the finished seed-$42$ run at the same step and weights. The $450/720$ and $900/360$ rows use a forecast trained and evaluated at $T_h=60\,\mathrm{s}$. The $1800/720$ row evaluates the $300\,\mathrm{s}$ mainline weights at $T_h=60\,\mathrm{s}$. Selective-Rerouter keeps the lowest civilian mean and the lowest object mean on every demand, and the most object deliveries on the three heaviest demands. It is short of a full delivery at $1800/1440$ and at both $2700$-civilian demands.

\begin{table}[!t]
\fontsize{6}{7}\selectfont
\setlength{\tabcolsep}{0.6pt}
\renewcommand{\arraystretch}{0.85}
\caption{Baseline comparison on eight mode-2 demands. Times are in seconds. Each row is one method. The two column groups are two demands. Bold is the best value in that block: the lowest time, or the highest $D_{\mathrm{osc}}$ and $D_{\mathrm{osc}}/M$. Underline is the second-best distinct value. A tie is marked on every such row.}
\label{tab:baseline900}
\begin{tabular*}{\columnwidth}{@{\extracolsep{\fill}}lrrrrr@{\hspace{2pt}}rrrrr@{}}
\toprule
& \multicolumn{5}{c}{$450/720$} & \multicolumn{5}{c}{$1800/720$} \\
\cmidrule(lr){2-6}\cmidrule(lr){7-11}
Method & \makecell[r]{$\bar{t}_{\mathrm{civ}}$\\(s)} & \makecell[r]{$\bar{T}_{\mathrm{obj}}$\\(s)} & \makecell[r]{$T_{\mathrm{obj}}^{\max}$\\(s)} & $D_{\mathrm{osc}}$ & \makecell[r]{\%} & \makecell[r]{$\bar{t}_{\mathrm{civ}}$\\(s)} & \makecell[r]{$\bar{T}_{\mathrm{obj}}$\\(s)} & \makecell[r]{$T_{\mathrm{obj}}^{\max}$\\(s)} & $D_{\mathrm{osc}}$ & \makecell[r]{\%} \\
\midrule
Proposed & \textbf{1646.2} & \textbf{1008.2} & \textbf{1874} & \textbf{720} & \textbf{100.00} & \textbf{1014.2} & \textbf{1098.8} & \rkii{1982} & \textbf{720} & \textbf{100.00} \\
NRR & $2041.8$ & $1175.8$ & $2758$ & \rkii{719} & \rkii{99.86} & $1550.5$ & $1572.4$ & $3307$ & $704$ & $97.78$ \\
ReFOCUS+ & $1861.6$ & $1114.3$ & $2212$ & \textbf{720} & \textbf{100.00} & $1282.9$ & $1280.8$ & $2208$ & \textbf{720} & \textbf{100.00} \\
FCR & $1925.7$ & $1128.8$ & $2310$ & \textbf{720} & \textbf{100.00} & $2376.1$ & $2291.2$ & $4610$ & $687$ & $95.42$ \\
CCRGS & $1891.4$ & $1134.5$ & $2157$ & \textbf{720} & \textbf{100.00} & $1334.0$ & $1422.3$ & $2477$ & $717$ & $99.58$ \\
Du-GAQ & $1842.7$ & $1136.7$ & $2219$ & \textbf{720} & \textbf{100.00} & $1270.1$ & $1349.7$ & $2416$ & \rkii{719} & \rkii{99.86} \\
CAIE-TT-Scoped & \rkii{1798.9} & \rkii{1087.8} & \rkii{2011} & \textbf{720} & \textbf{100.00} & \rkii{1049.6} & \rkii{1126.9} & \textbf{1940} & \textbf{720} & \textbf{100.00} \\
\midrule
& \multicolumn{5}{c}{$900/360$} & \multicolumn{5}{c}{$1800/1440$} \\
\cmidrule(lr){2-6}\cmidrule(lr){7-11}
Proposed & \textbf{691.6} & \textbf{675.4} & \rkii{1248} & \textbf{360} & \textbf{100.00} & \textbf{2378.5} & \textbf{2042.8} & \textbf{3612} & \textbf{1439} & \textbf{99.93} \\
NRR & $885.8$ & $792.4$ & $1567$ & \textbf{360} & \textbf{100.00} & $8543.8$ & $6698.6$ & $36337$ & $1260$ & $87.50$ \\
ReFOCUS+ & $820.7$ & $771.1$ & $1358$ & \textbf{360} & \textbf{100.00} & $15687.5$ & $10278.4$ & $101767$ & $1208$ & $83.89$ \\
FCR & $910.3$ & $772.8$ & $1364$ & \textbf{360} & \textbf{100.00} & $11546.4$ & $9709.0$ & $34068$ & $1251$ & $86.88$ \\
CCRGS & $787.2$ & $762.0$ & $1359$ & \textbf{360} & \textbf{100.00} & $4364.8$ & $3704.0$ & $6808$ & $1383$ & $96.04$ \\
Du-GAQ & \rkii{742.8} & $756.6$ & $1289$ & \textbf{360} & \textbf{100.00} & $4678.1$ & $3841.3$ & $8418$ & $1336$ & $92.78$ \\
CAIE-TT-Scoped & $779.2$ & \rkii{754.6} & \textbf{1217} & \textbf{360} & \textbf{100.00} & \rkii{2665.7} & \rkii{2180.9} & \rkii{4028} & \rkii{1432} & \rkii{99.44} \\
\midrule
& \multicolumn{5}{c}{$900/720$} & \multicolumn{5}{c}{$2700/1440$} \\
\cmidrule(lr){2-6}\cmidrule(lr){7-11}
Proposed & \textbf{1366.7} & \textbf{1040.2} & \rkii{1958} & \textbf{720} & \textbf{100.00} & \textbf{8901.6} & \textbf{10506.9} & \textbf{19348} & \textbf{1308} & \textbf{90.83} \\
NRR & $1667.1$ & $1181.8$ & $2255$ & $718$ & $99.72$ & $25033.9$ & $21288.2$ & $101312$ & $996$ & $69.17$ \\
ReFOCUS+ & $1928.1$ & $1373.9$ & $2679$ & $715$ & $99.31$ & $20865.6$ & $20285.2$ & $101972$ & $979$ & $67.99$ \\
FCR & $1583.0$ & $1144.8$ & $2048$ & \textbf{720} & \textbf{100.00} & $18887.5$ & $24899.8$ & $75558$ & \rkii{1172} & \rkii{81.39} \\
CCRGS & $1525.0$ & $1157.8$ & $1966$ & \textbf{720} & \textbf{100.00} & $17691.3$ & $18160.2$ & $101838$ & $1046$ & $72.64$ \\
Du-GAQ & $1511.7$ & $1162.2$ & $2077$ & \rkii{719} & \rkii{99.86} & $18264.8$ & \rkii{16473.7} & $101355$ & $995$ & $69.10$ \\
CAIE-TT-Scoped & \rkii{1394.5} & \rkii{1078.0} & \textbf{1823} & \textbf{720} & \textbf{100.00} & \rkii{12510.9} & $17994.5$ & \rkii{70967} & $1152$ & $80.00$ \\
\midrule
& \multicolumn{5}{c}{$900/1080$} & \multicolumn{5}{c}{$2700/2160$} \\
\cmidrule(lr){2-6}\cmidrule(lr){7-11}
Proposed & \textbf{1967.9} & \textbf{1378.2} & \textbf{2532} & \textbf{1080} & \textbf{100.00} & \textbf{18462.7} & \textbf{10494.5} & \textbf{101342} & \textbf{1754} & \textbf{81.20} \\
NRR & $2274.5$ & $1502.5$ & $3088$ & $1071$ & $99.17$ & $66661.7$ & $53889.1$ & $102327$ & $1459$ & $67.55$ \\
ReFOCUS+ & $2168.3$ & $1499.7$ & $2693$ & $1072$ & $99.26$ & $59176.3$ & $44173.3$ & $102302$ & $1464$ & $67.78$ \\
FCR & $2355.6$ & $1600.9$ & $3152$ & $1066$ & $98.70$ & $59443.9$ & $47069.9$ & $103167$ & $1383$ & $64.03$ \\
CCRGS & $2215.5$ & $1534.8$ & $2970$ & $1073$ & $99.35$ & $55702.2$ & $43870.0$ & \rkii{101378} & $1381$ & $63.94$ \\
Du-GAQ & $2164.9$ & $1508.9$ & $2792$ & \rkii{1074} & \rkii{99.44} & $54213.0$ & $44022.9$ & $101404$ & $1449$ & $67.08$ \\
CAIE-TT-Scoped & \rkii{1990.4} & \rkii{1410.4} & \rkii{2627} & \textbf{1080} & \textbf{100.00} & \rkii{24759.5} & \rkii{13640.0} & $102885$ & \rkii{1664} & \rkii{77.04} \\
\bottomrule
\end{tabular*}
\end{table}

\subsection{Interpretation and Scope of Experimental Results}
\label{sec:exp-scope}
Section~\ref{sec:scope} defines the problem scope. Tables~\ref{tab:hybrid400}--\ref{tab:baseline900} are one-run SUMO comparisons with cost-announcement-minimizing gateway selection and a frozen forwarder. Corner counts in Tables~\ref{tab:zoneexit} and~\ref{tab:zoneproto} are intermediate checks and do not replace the end-to-end scores of Section~\ref{sec:metrics}. The simulator omits temperature and stamp verification. Section~\ref{sec:conclusion} records the remaining limits.
\setcounter{secnumdepth}{3}

\section{Conclusion}
\label{sec:conclusion}
\label{sec:conclusion-intact}
The primary result is the Selective-Rerouter for a hypothetical post-disaster scenario whose road network is partitioned into DTN subregions. On intact streets it replans only congestion-related vehicles. Trials on the intact Tainan road-network topology attribute the main travel-time gain to the live--forecast blend~\cite{WangYang2025LSTAN}. Four-corner trials score the Selective-Rerouter in our co-simulator with the DTN-UAV Forwarder held fixed (Table~\ref{tab:baseline900}).

\textbf{Threats to validity.}
\label{subsec:limitations}
Corner-check counts in Section~\ref{sec:sub1val} do not enter Table~\ref{tab:baseline900}. Every comparison is one SUMO run, with no outdoor flight, no packet loss, and fixed oscillator trajectories. Civilians follow the written route. Intact results use the Tainan road-network topology with no injected partition. The four-corner setup is one artificially injected partition of that topology and does not represent a real-world disaster at this location.

\textbf{Future work.}
\begin{itemize}
\item Repeat Section~\ref{sec:sub1val} with a cost-announcement that moves with the intact fleet. The simulated cost-announcement used there is constant, so $T_{\mathrm{ref}}$ did not rename the gateway.
\item Compare the DTN-UAV Forwarder with another forwarder that obeys the same cost-announcement, on a lossy contact model.
\item Repeat the Selective-Rerouter comparison on a second partitioned map.
\end{itemize}

\bibliographystyle{ieeetr}
\bibliography{references}

\end{document}